\documentclass[10pt,journal,compsoc]{IEEEtran}
\usepackage{subfiles}

\ifCLASSOPTIONcompsoc
  \usepackage[nocompress]{cite}
\else
  \usepackage{cite}
\fi
\ifCLASSINFOpdf
  \usepackage[pdftex]{graphicx}
  \graphicspath{{figs/}}
\else
\fi
\usepackage{amsmath}
\usepackage{algorithm}
\usepackage{algpseudocode}
\newcommand*\Let[2]{\State #1 $\gets$ #2}
\algrenewcommand\algorithmicrequire{\textbf{Input:}}
\algrenewcommand\algorithmicensure{\textbf{Output:}}

\usepackage{array}

\ifCLASSOPTIONcompsoc
 \usepackage[caption=false,font=footnotesize,labelfont=sf,textfont=sf]{subfig}
\else
 \usepackage[caption=false,font=footnotesize]{subfig}
\fi
\usepackage{url}

\usepackage{booktabs}
\usepackage{multicol}
\usepackage{multirow}
\usepackage[inline]{enumitem}
\usepackage{amsmath,amsfonts,amssymb,bm}
\usepackage{tikz}
\usetikzlibrary{math}
\usepackage{pgfplots}
\usepackage{pgfplotstable}
\usepgfplotslibrary{groupplots}
\usetikzlibrary{shapes.geometric}
\usepackage{siunitx}
\usepackage[para]{threeparttable}
\usepackage{mathtools}
\usepackage{etoolbox}
\newrobustcmd\ubold{\DeclareFontSeriesDefault[rm]{bf}{b}\bfseries}
\usepackage[normalem]{ulem}
\robustify\bfseries
\robustify\uline
\robustify\tnote
\def\Uline#1{#1\llap{\uline{\phantom{#1}}}}
\pgfplotsset{compat=1.18}
\usepackage{amsmath,amsfonts,bm}
\usepackage{xspace}
\makeatletter
\DeclareRobustCommand\onedot{\futurelet\@let@token\@onedot}
\def\@onedot{\ifx\@let@token.\else.\null\fi\xspace}

\def\eg{\emph{e.g}\onedot}

\def\etc{\emph{etc}\onedot} 
\def\wrt{w.r.t\onedot} 
\def\etal{\emph{et al}\onedot}
\makeatother

\def\eqref#1{equation~\ref{#1}}

\def\1{\bm{1}}

\def\rmA{{\mathbf{A}}}
\def\rmB{{\mathbf{B}}}

\def\rmD{{\mathbf{D}}}
\def\rmE{{\mathbf{E}}}

\def\rmH{{\mathbf{H}}}
\def\rmI{{\mathbf{I}}}

\def\rmK{{\mathbf{K}}}

\def\rmP{{\mathbf{P}}}
\def\rmQ{{\mathbf{Q}}}

\def\rmV{{\mathbf{V}}}
\def\rmW{{\mathbf{W}}}

\def\rmY{{\mathbf{Y}}}

\def\vb{{\bm{b}}}

\def\vd{{\bm{d}}}

\def\vk{{\bm{k}}}

\def\vn{{\bm{n}}}

\def\vp{{\bm{p}}}
\def\vq{{\bm{q}}}
\def\vr{{\bm{r}}}

\def\vv{{\bm{v}}}

\DeclareMathAlphabet{\mathsfit}{\encodingdefault}{\sfdefault}{m}{sl}
\SetMathAlphabet{\mathsfit}{bold}{\encodingdefault}{\sfdefault}{bx}{n}

\newcommand{\R}{\mathbb{R}}

\DeclareMathSymbol{\shortminus}{\mathbin}{AMSa}{"39}
\begin{document}
%
\title{RepSGG: Novel Representations of Entities and Relationships for Scene Graph Generation}
%
%
%
%

\author{Hengyue Liu,
\thanks{hliu087@ucr.edu} 
        Bir Bhanu~\IEEEmembership{Life~Fellow,~IEEE}%
        
}

%
%

\markboth{}%
{Shell \MakeLowercase{\textit{et al.}}: Bare Demo of IEEEtran.cls for Computer Society Journals}
%



\IEEEtitleabstractindextext{%
\begin{abstract}
Scene Graph Generation (SGG) has achieved significant progress recently. However, most previous works rely heavily on fixed-size entity representations based on bounding box proposals, anchors, or learnable queries. As each representation's cardinality has different trade-offs between performance and computation overhead, extracting highly representative features efficiently and dynamically is both challenging and crucial for SGG. In this work, a novel architecture called RepSGG is proposed to address the aforementioned challenges, formulating a subject as queries, an object as keys, and their relationship as the maximum attention weight between pairwise queries and keys. With more fine-grained and flexible representation power for entities and relationships, RepSGG learns to sample semantically discriminative and representative points for relationship inference. Moreover, the long-tailed distribution also poses a significant challenge for generalization of SGG. A run-time performance-guided logit adjustment (PGLA) strategy is proposed such that the relationship logits are modified via affine transformations based on run-time performance during training. This strategy encourages a more balanced performance between dominant and rare classes. Experimental results show that RepSGG achieves the state-of-the-art or comparable performance on the Visual Genome and Open Images V6 datasets with fast inference speed, demonstrating the efficacy and efficiency of the proposed methods.
\end{abstract}

}

\maketitle

\IEEEdisplaynontitleabstractindextext

%
\IEEEpeerreviewmaketitle

\IEEEraisesectionheading{\section{Introduction}\label{sec:introduction}}

%
%
%
%
\IEEEPARstart{T}{o} understand a scene, it is important to infer underlying properties of entities and the relationships between them. 
For a computer vision system to explicitly represent and reason about the detailed semantics, Johnson~\cite{johnson2015image}~\etal adopt and formalize scene graphs from computer graphics community. 
A scene graph is an explicit graph representation for modeling a visual scene, where entities are the nodes, and pairwise relationships are represented as edges. 
A relationship between two entities is denoted as a triplet of \texttt{<}\texttt{subject}, \texttt{predicate}, \texttt{object>}. 
In the context of this paper, the term ``entity'' denotes an instance of an object, while the term ``object'' specifically indicates an entity with semantic significance. 
Serving as a powerful representation, scene graph enables many down-stream high-level reasoning tasks such as image captioning~\cite{yang2019auto,yao2018exploring}, image retrieval~\cite{johnson2018image,johnson2015image}, Visual Question answering~\cite{hudson2019gqa,teney2017graph,li2022inner} and image generation~\cite{johnson2018image,xu2018attngan}. 
Since SGG is built upon object detection where many off-the-shelf detectors~\cite{girshick2015fast,ren2015faster,newell2016stacked,he2017mask,zhou2019objects,tian2020fcos} can be used, limited attention has been directed towards the investigation of better feature representations for entities and relationships. 
Currently, there are mainly three types of entity visual feature representations: 
\begin{figure}[t]
\centering
\includegraphics[width=\linewidth]{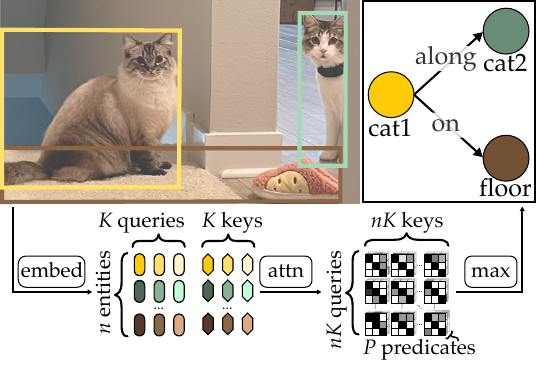}
\caption{Illustration of RepSGG. For $n$ detected entities, each entity is represented by $K$ subject queries and $K$ object keys. The attention weights between queries and keys are projected as the predicate classification scores in the shape of $P \times nK \times nK $, where $P$ is the number of predicates in a dataset. The final predicate classification is reduced to the shape of $P \times n \times n $ by max-pooling, and top predictions are collected as the scene graph. $K = 3$ and $n = 3$ in the example.}
\label{fig:teaser}
\end{figure}
\begin{enumerate}
  \item \emph{Box-based}: spatially-pooled features extracted from bounding boxes. Most SGG methods~\cite{xu2017scene,tang2020unbiased} are based on R-CNN detectors~\cite{girshick2015fast,ren2015faster,he2017mask} which use RoIPooling~\cite{girshick2015fast} or RoIAlign~\cite{he2017mask} for feature extraction. 
  \item \emph{Point-based}: single-pixel features extracted from bounding box centers. These methods~\cite{newell2017pixels,liu2021fully,Adaimi_2023_WACV} utilize anchor-free detectors~\cite{newell2016stacked,zhou2019objects,tian2020fcos} to ground entities and relationships in a regression fashion. 
  \item \emph{Query-based}: fixed-size learnable embeddings. These methods~\cite{li2022sgtr,desai2022single,cong2023reltr} build upon DETR~\cite{carion2020end} where message passing and matching are performed between entity and relationship embeddings.
\end{enumerate}

Each type has its own merits and drawbacks. While preserving the spatial appearance of entities, box-based features are computationally expensive and use more memory. Point-based features are often regressed and extracted at the center of an entity's bounding box (referred to as entity center for the rest of the paper). Although such models achieve fast inference speed, their performance is relatively low due to the fact that point-based features are less semantically meaningful with limited cardinality. Query-based representation relies on a fixed number of learnable embeddings for decoding entities or relations, which considerably simplifies the SGG tasks. However, it suffers from training difficulties and lower performance compared to box-based methods. 

Another issue of box-based, point-based and query-based entity representations lies in their predetermined granularities. Box-based features could be coarse and redundant, while point-based and query-based features are insufficient to represent entities with different semantics. For example, if there are two relation for the same entity \texttt{person}, \texttt{<person, eating, pizza>} and \texttt{<person, on, street>}, it is important to capture the context around the mouth and hands of the person for the first triplet, and around the feet and street for the second one. Box-based representation keeps the information of the entity \texttt{person}, but loses the fine-grained features around the mouth, hands and feet due to the low pooling resolution (\eg $7 \times 7 $). The same issue will arise for point-based and query-based representations. Increasing the pooling resolution, feature dimension, or number of queries could help, but the computation complexity will increase dramatically.

Regarding the relation representations, most works use compositional contextual features to perform predicate classification. Few researchers~\cite{Adaimi_2023_WACV,liu2021fully} have explored different ways to represent relations in a regression fashion. These methods represent entities as keypoints (\eg, entity centers), and relationships as points~\cite{newell2017pixels} or vectors~\cite{liu2021fully,Adaimi_2023_WACV}. By regressing and grounding entities and relations geometrically, these methods achieve faster inference speed which is useful for down-stream tasks. However, regression with handcrafted targets~\cite{liu2021fully,Adaimi_2023_WACV} is deficient, especially on sparsely-annotated datasets~\cite{krishna2017visual}. Consequently, the performance of regression is inferior compared to predicate classification.

In this work, we seek new insights into alternative representations of entities and relationships for scene graph generation. 
We propose a novel architecture called RepSGG that is built upon the FCOS~\cite{tian2020fcos} entity detector with a specialised transformer-based~\cite{vaswani2017attention,carion2020end,zhu2021deformable} relationship decoder. 
As shown in Fig.~\ref{fig:teaser}, an entity class is represented with a set of learnable subject and object embeddings, named rep-embeddings, to learn diverse semantics.
The subject and object embeddings are progressively augmented by dynamically sampled visual features and inter-embedding attentions through decoder layers.
First, each embedding is updated by visual features sampled at semantic-dependent representative points (rep-points); then, a two-way cross-attention (subject-to-object and vice-versa) is performed to update both subject and object embeddings.
For relationship prediction, subject and object embeddings are treated as queries and keys, with relationships quantified as the projected attention weights between these queries and keys.


Besides the proposed architecture, we also investigate the long-tailed problem in SGG. 
Inspired by logit adjustment~\cite{menon2021longtail}, we propose a run-time performance-guided logit adjustment (PGLA) to achieve per-instance label-dependent loss modification. 
We measure and update the performance (\eg, recall or precision) of predicate predictions per mini-batch, per iteration during training. 
Instead of adding a bias term to logits as in \cite{menon2021longtail}, we perform the affine transformation on the logits.
We also measure run-time logit differences among predicates, named confusion logits, to further enlarge inter-class logit margins.
For each predicate, the among of the adjustment will be determined by its frequency in the training set, run-time performance, and confusion logits. 

The contributions of this paper are:
\begin{enumerate}
\item Significantly different from most existing SGG approaches, RepSGG introduces a novel SGG paradigm in which entities are expressed as queries and keys, and relationships are represented as their attention weights. It explores a natural approach of capturing visual and semantic features progressively, and encapsulating relationships as attention weights which encode the edge confidence and directionality effectively.

\item A run-time performance-guided logit adjustment (PGLA) strategy is proposed to mitigate the long-tailed problem. PGLA is a simple, yet effective, model-agnostic, and cost-free solution that achieves a more balanced performance on unbalanced data. The choice of loss (\eg, binary cross-entropy or cross-entropy) and performance metric (\eg, recall or precision) are task-dependent, and this adaptability can be extended to a range of settings and tasks.  

\item Extensive experiments on the Visual Genome and Open Images V6 datasets demonstrate the effectiveness of the proposed approach. Beyond standard SGG metrics, we also report the zero-shot mean recall (zs-mR) on our method and several state-of-the-art methods. RepSGG exhibits superior robustness and generalization capabilities on out-of-distribution data.

\end{enumerate}

The remainder of the paper is structured as follows: Section~\ref{sec:relatedwork} provides a literature review of related work in scene graph generation. Section~\ref{sec:approach} outlines the proposed approach. Section~\ref{sec:experiments} discusses experimental results and ablation studies. Section~\ref{sec:lim} presents the limitations and future work. Section~\ref{sec:conclusions} concludes this paper.


\section{Related Work}\label{sec:relatedwork}

The task of scene graph generation involves numerous aspects, and our focus lies on entity and relation representations, as well as the long-tailed problem.

\subsection{Feature Representations}

As discussed in Section~\ref{sec:introduction}, the entity representation is either box-based, point-based, or query-based.
Traditional SGG methods~\cite{xu2017scene,zellers2018neural,tang2019learning} utilize a pre-trained detector~\cite{ren2015faster,he2017mask} to extract a set of entity bounding boxes and their corresponding feature maps via feature pooling (RoIPool~\cite{ren2015faster} or RoIAlign~\cite{he2017mask}). The visual features for each entity, commonly referred to as appearance features, are represented as a tensor of shape $c \times h \times w$, where $c$ is the number of channels of the feature, and $h \times w$ is the feature spatial size. Those features are used to construct visual context for predicate classification. To incorporate the relative position between entities, researchers use the geometric layout encoding~\cite{woo2018linknet}, union of bounding boxes mask encoding~\cite{dai2017detecting}, or geometric constraints~\cite{wang2019exploring} for a better visual representation. 

One-stage anchor-free entity detectors~\cite{law2018cornernet,zhou2019objects,tian2020fcos} have recently gained popularity due to their simplicity and efficiency. In these works, entities are directly regressed at pixels in feature maps, where the entity center or corners are selected as the ground-truth targets. Instead of pooling features of various shapes, extracting features at multiple pixels is much faster and consumes less memory. Several works~\cite{newell2017pixels,liu2021fully,Adaimi_2023_WACV} explore such point-based entity representation for SGG. Pixel2Graph~\cite{newell2017pixels} grounds edges at the midpoints between the bounding box centers of subjects and objects (referred to as subject and object centers for the rest of the paper). FCSGG~\cite{liu2021fully} uses relation affinity fields to encode the relations as 2D vectors ``flow'' from the subject to object centers. CoRF~\cite{Adaimi_2023_WACV} extends the concept of fields by composing more regression targets per pixel. 

Recent works in scene graph generation have explored transformer-based models to improve the performance. Several works~\cite{tang2020unbiased,lu2021context,dhingra2021bgt} start to replace the RNN-based context decoders~\cite{zellers2018neural,tang2019learning} with multi-head self-attention~\cite{vaswani2017attention}. Subsequently, other works~\cite{dong2021visual,teng2022structured,chen2022reltransformer} explore ways to construct subject, object and predicate queries with variants of transformers. With the success of DETR~\cite{carion2020end}, more works~\cite{li2022sgtr,desai2022single,cong2023reltr} study the query-based representations of entities and relations. For DETR-like approaches, there are a fixed number of learnable entity and predicate queries, which will be decoded as output triplets in an end-to-end manner.  

The strengths and weaknesses of different types of entity representation vary based on their granularity and flexibility. For example, box-based features are extracted via RoIAlign~\cite{he2017mask} with a fixed shape of $d \times h \times w$, such as $256 \times 7 \times 7$ for SGG tasks. 
Although preserving entities' spatial configuration, box-based features may lose semantic details due to the pooling operation.
Furthermore, the fact that features are pooled into the same shape regardless of actual sizes of entities may result in the loss of semantic details in relationship inference. 
Another drawback of the box-based representation is that it is computationally expensive to compute $\mathcal{O}(n^2)$ relationships for $n$ entity proposals. Sampling candidate entities and relationships is commonly used during training. 
On the other hand, the point-based methods significantly reduce the computational cost by using features of shape $ d \times 1$. By reformulating the SGG in a per-pixel regression fashion, point-based methods~\cite{liu2021fully,Adaimi_2023_WACV} achieve much faster inference speed. However, the performance is relatively lower due to the coarse entity representations and handcrafted relationship targets. 
The query-based entity representation provides a way to perform object detection and SGG in an end-to-end manner. It exhibits greater capability in capturing semantics compared to convolutional regression, achieving better performance than point-based methods.
Nevertheless, challenges arise for query-based methods due to factors like feature cardinality, the constraint of a fixed number of learnable queries, and increased complexity in both model design and post-processing.
It is also difficult for query-based methods to perform predicate classification and scene graph classification due to the end-to-end prediction manner.

This work proposes a novel entity representation by using a set of semantically representative embeddings, which are augmented by visually representative points (rep-points)~\cite{dai2017deformable,yang2019reppoints} progressively. 
Different sets of rep-points are sampled dynamically \wrt entity reference points (centers) to update subject and object embeddings, respectively. 
Cross-attention between subject and object embeddings are also performed to achieve message passing. 
This approach allows for each entity instance to be represented as distinct queries and keys, which is more flexible than box-based representation and more fine-grained than point-based and query-based representations. 
Additionally, the proposed method eliminates the need for composite or predicate queries~\cite{dong2021visual,cong2023reltr,li2022sgtr}, as the predicate of a relationship can be computed as the multi-head attention weights between the subjects (as queries) and objects (as keys). 
Unlike triplet classification, modeling relationships as attention weights preserves the directional information among subjects and objects, and captures more semantics.

\subsection{Long-tailed Distributions}

Long-tailed data distribution has been a key challenge in visual recognition~\cite{zhang2023deep}, and it has been addressed in the recent literature on SGG~\cite{chang2021comprehensive}. In order to tackle this problem, various approaches have been proposed, such as data re-sampling~\cite{li2021bipartite,desai2021learning,li2022devil,zhang2022fine}, de-biasing~\cite{wang2020tackling,tang2020unbiased,chiou2021recovering,guo2021general,he2021learning}, and loss modification~\cite{gkanatsios2020saturation,knyazev2020graphdensity,lin2020gps,yan2020pcpl,suhail2021energy,li2022ppdl,lyu2022fine}. De-biasing methods require pre-trained biased models for initialization and then finetune the model. Loss modification methods generally assign a weight vector to the cross-entropy loss for predicate classification, with higher weights to tail classes and lower weights to head classes. 

In this work, instead of re-weighting the loss function, another type of approach is applied by directly modifying the classification logits~\cite{menon2021longtail,Zhang_2021_CVPR,wei2022mitigating,chen2022resistance}. A run-time performance-guided logit adjustment strategy is presented which offers a dynamic and effective control over the relative contributions of labels in the loss.

\section{Technical Approach}\label{sec:approach}

\begin{figure*}[t]
\centering
\includegraphics[width=\linewidth]{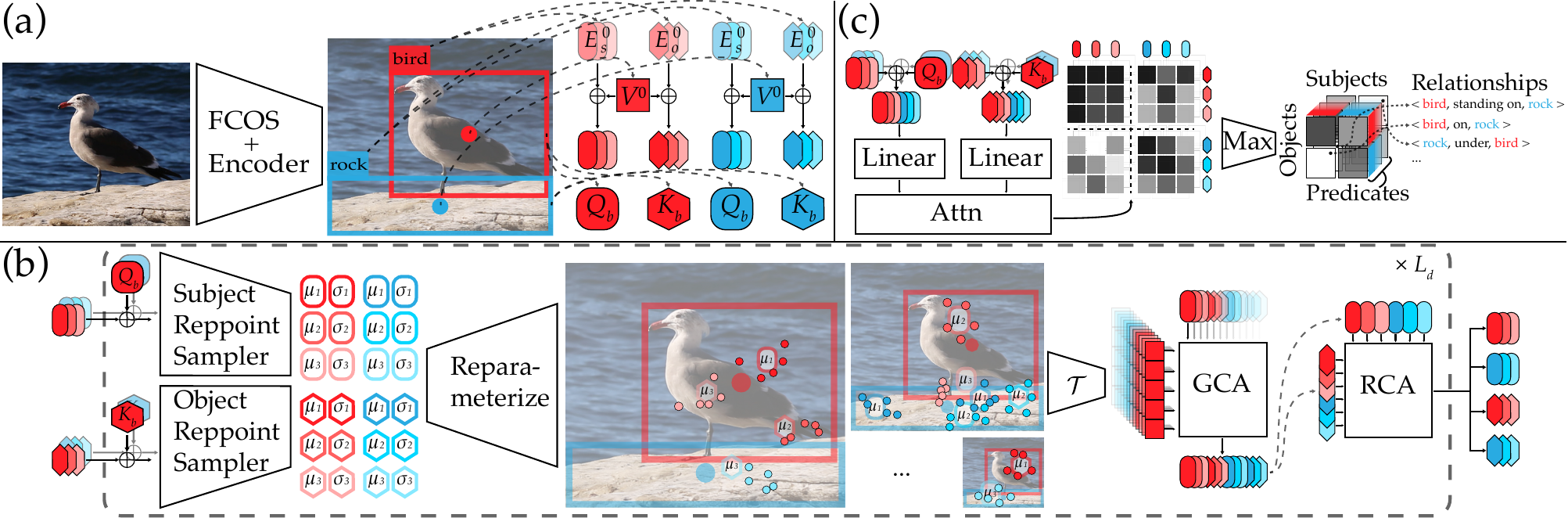}
\caption{An illustration of the pipeline of RepSGG: (a) initial queries (subjects) and keys (objects) are generated from detected entities and encoded features; (b) queries and keys are updated through the decoder; (c) the relationship confidence scores are computed as the maximum attention weights between queries and keys.}
\label{fig:arch}
\end{figure*}

In this section, we first provide the preliminaries of modeling entity detection in a per-pixel prediction fashion. Subsequently, we introduce the RepSGG architecture, consisting of an entity detector, encoder, decoder, and relationship output layer. An illustration of RepSGG is shown in Fig.~\ref{fig:arch}. Finally, we present several training strategies, including PGLA, addressing the challenges posed by long-tailed distributions and sparsely annotated data. 

\subsection{Entity Detection}

Our model is built upon a one-stage anchor-free detector, namely FCOS~\cite{tian2020fcos}. Different from commonly used anchor-based R-CNN approaches for generating object proposals and features, entity detections are decoded from regressed dense features. More specifically, an input image $\rmI \in \R^{H^0 \times W^0 \times 3}$ will go through a backbone CNN (\eg, ResNet-50~\cite{he2016deep}) followed by a feature pyramid network (FPN)~\cite{lin2017feature}, generating 5 scale levels of visual feature maps. Following FCOS~\cite{tian2020fcos}, feature maps of different levels are responsible for entities of different sizes. Shared detection heads are used for predicting dense feature maps that provide entity classification, bounding box regression, and center-ness. The entity detections can be decoded from the predicted feature maps as $ \mathcal{B}:\{ b^i\}_{i=1}^n $, where $ b^i = (x_0^i, y_0^i, x_1^i, y_1^i, z^i, c^i)$, $ (x_0^i, y_0^i) $ and $ (x_1^i, y_1^i) $ denote the coordinates of the top-left and bottom-right corners of the bounding box, $z^i \in \{0,1,2,3,4 \}$ is the scale level at which the detection is decoded, $c^i \in \{0, ..., C-1 \}$ is the predicted entity label for a dataset containing $C$ classes, and $n$ is the number of detected entities. Since the positive training targets are defined inside the center area of a bounding box, the center of the bounding box is considered as the reference point of an entity where the class and bounding box are regressed.

\subsection{Encoder} \label{sec:encoder}

To further encourage information exchange among different scales and prepare for the relationship decoder, a deformable transformer encoder~\cite{zhu2021deformable}, consisting of $L_e$ encoder layers, is applied on the FPN features without changing their shapes. The output features are considered as the visual features of the image. Following~\cite{zhu2021deformable}, the 2D positional embeddings~\cite{vaswani2017attention} are generated and added with a learnable scale-level embedding. Note that the visual features and positional embeddings are multi-scale features. To merge levels of features, they are resized via bilinear interpolation to the shape of the largest feature map respectively. The interpolated features are then stacked along the level dimension, producing the visual features $\rmV \in \R^{5 \times H \times W \times d} $ and positional embeddings $\rmP\rmE \in \R^{5 \times H \times W \times d}$, where $ H = \lfloor H^0 / 8 \rfloor $, $ W = \lfloor W^0 / 8 \rfloor $ and $d = 256$.

Our core idea is to represent an entity by a set of distinct representative features in the semantic space. Concretely, we construct subject representative embeddings (rep-embeddings) $\rmE_s$ and object rep-embeddings $\rmE_o$ of shape $ C \times K \times d$ , where $K$ is the number of entity's class-inherent embeddings.
Rep-embeddings, with $2K$ distinct learnable embeddings per entity class, provide semantically rich yet discriminative representations of entities.

\subsection{Decoder}

The decoder is composed of a stack of $L_d$ decoder layers, where the initial inputs are formed by fusing subject features with embeddings, as well as object features with embeddings. 
The outputs are decoded subject and object features of the same shape as inputs. 
Each layer has a rep-point sampler, two group cross-attention layers, and a two-way relational cross-attention layer. 
To simplify the terms and make them compatible with the attention mechanism, we refer to the subject embeddings as \emph{queries}, and object embeddings as \emph{keys} for the rest of the paper.

\subsubsection{Initial Queries and Keys}

The entity-related features are firstly sampled to create the initial queries and keys.
Let the reference point $ \vp \in [ 0, 1 ]^3$ be the normalized coordinates, where $ (0,0,0) $ and $ (1,1,1) $ indicate the top-left corner at the lowest scale level, and bottom-right corner at the highest scale level of the features, then for $n \times m$ reference points $ \rmP =  \{ \vp^{i,1}, \ldots, \vp^{i,m} \}_{i=1}^n $, the point sampler function is defined as
\begin{equation}
\mathcal{T}(\;\cdot\;, \rmP): \R^{5 \times h \times w \times d} \times \R^{n \times m \times 3} \rightarrow \R^{n \times m \times d},
\label{eq:sampler}
\end{equation}
which is achieved by bilinear interpolation.
To prepare semantic-agnostic entity features, the normalized centers of bounding boxes $ \rmP^0 \in \R^{n \times 1 \times 3} $ are used as the reference points derived from the entity detections $\mathcal{B}$. The point sampler is applied on $\rmV$ and $\rmP\rmE$ to get the corresponding features at the entities' reference points as:
\begin{equation}
\begin{aligned}
\rmV^0 & =  \mathcal{T}(\rmV,  \rmP^0) \\
\rmP \rmE^0 & =  \mathcal{T}(\rmP\rmE, \rmP^0) \\
\rmP^0 & = \left\{ \left( \frac{x_0^i + x_1^i}{2W}, \frac{y_0^i + y_1^i}{2H}, \frac{z^i}{4} \right) \right\}_{i=1}^n.
\end{aligned}
\label{eq:f0}
\end{equation}
For assigning semantics to entities, the subject and object embeddings are gathered from the corresponding indices of predicted entity labels $\{ c^i\}_{i=1}^n $ as $\rmE_s^0 \in \R^{n \times K \times d}$ and $\rmE_o^0 \in \R^{n \times K \times d}$. The subject and object embeddings are class-dependent learnable parameters which capture the semantics of an entity class being the subject and object.
Subject queries $\rmQ^0$ and object keys $ \rmK^0 $ are then constructed by adding the corresponding embeddings with the visual features $\rmV^0$:
\begin{equation}
\begin{aligned}
\rmQ^0 & = \rmE_s^0 + \rmV^0 \\
\rmK^0 & =  \rmE_o^0 + \rmV^0.
\end{aligned}
\label{eq:qk}
\end{equation}
By merging instance-specific visual features with class-specific embeddings, the queries (or keys) retain semantic similarities within their entity classes while also diversifying the instance-wise entity representations.

The bounding box also plays a crucial role in determining the spatial relationship between entities.
Hence, the bounding box coordinates are mapped into embeddings. 
First, the positional embeddings $\rmP\rmE$ of the top-left and bottom-right corners are sampled via (\ref{eq:sampler}). 
Two learnable embeddings indicating ``top-left corner'' and ``bottom-right corner'' are added with the corresponding corners' positional embeddings, respectively. 
The two corner embeddings are then concatenated and fed to a fully-connected layer, resulting in the box embeddings. 
Lastly, two learnable embeddings are added with the box embeddings to construct subject and object box embeddings respectively, denoted as $\rmQ_{b} \in \R^{n \times d}$ and $\rmK_{b} \in \R^{n \times d}$.

\subsubsection{Rep-point Sampler} \label{sec:reppoint_sampler}
For rep-embeddings to capture more visual context, we propose a dynamic approach for sampling features from representative points (rep-points) and message passing via attentions.
Two rep-point samplers are implemented to sample subject and object rep-points, since queries and keys serve distinct roles in conveying subject and object semantics respectively. Specifically, a multi-layer perceptron (MLP) is used as the sampler to predict the points of interest for each entity. To achieve semantic-specific sampling, the MLP weights are split into $K$ groups (in practice, group 1D convolution is used), and $k$-th group is applied on the $k$-th slice of the inputs (queries or keys) along the $K$ dimension of rep-embeddings. 
To further increase the diversity of sampling and prevent overfitting on few points, instead of directly predicting the coordinates, the distribution parameters of offsets \wrt the reference points are predicted, following the variational autoencoder (VAE) and reparameterization trick~\cite{kingma2013auto}. Let the input queries to the $l$-th layer be $\rmQ^{l-1}$, the subject rep-point offsets are defined as 
\begin{equation}
\begin{aligned}
 \Delta \rmP^l_s & = \bm{\mu}^l_s + \bm{\sigma}^l_s \odot \bm{\epsilon}, \\
 \bm{\mu}^l_s, \bm{\sigma}^l_s  & = \text{MLP}(\rmQ^{l-1} + \rmQ_{b})
 \end{aligned}
\end{equation}
where means $ \bm{\mu}^l_s \in \R^{n \times K \times 3} $ and standard deviations (stds) $ \bm{\sigma}^l_s \in \R^{n \times K \times 3} $ are the outputs of the subject rep-point sampler, $\odot$ is the element-wise product, and $ \bm{\epsilon} \sim \mathcal{N} (\bm{0}, \bm{I}_3)$.
To estimate robust parameters, $m$ points are randomly sampled per parameter during training, namely $ \bm{\epsilon} \in \R^{n \times K \times m \times 3} $.
Similarly, the object rep-point offsets are obtained as $ \Delta \rmP^l_o$. By adding the sampled rep-points offsets to the reference points, the sampled points are obtained as
\begin{equation}
\begin{aligned}
  \rmP^l_s & = \rmP^{l-1}_s + \Delta \rmP^l_s, \; l =1 \ldots L \\
  \rmP^l_o & = \rmP^{l-1}_o + \Delta \rmP^l_o, \; l =1 \ldots L,
 \end{aligned}
\end{equation}
where $\rmP^{0}_s = \rmP^{0}_o = \rmP^0$ with expanded shapes of $n \times 1 \times 1 \times 3$.
The offsets are accumulated through decoder layers from the original entity centers $\rmP^0$, so relevant features can be aggregated as the decoder layer goes deeper. Accordingly, rep-point visual features and positional embeddings are sampled via (\ref{eq:sampler}):
\begin{equation}
\begin{aligned}
 & \rmV^l_s =  \mathcal{T}(\rmV,  \rmP^l_s), \; \rmP\rmE^l_s = \mathcal{T}(\rmP\rmE, \rmP^l_s),\\
 & \rmV^l_o =  \mathcal{T}(\rmV,  \rmP^l_o), \; \rmP\rmE^l_o = \mathcal{T}(\rmP\rmE, \rmP^l_o).
\end{aligned}
\end{equation}

The rep-point sampler provides a probabilistic mapping from visual cues to the semantic space, which helps finding visually relevant features with semantic significance.
During inference, this stochastic behavior is transformed into a deterministic one by sampling within a range with a fixed step size. 
By default, rep-points are sampled within the ``$3\sigma$'' range with a step size of ``$\sigma$'' along the width, height, and scale dimensions in a combinatorial manner.
For a subject rep-point sampler for the $k$-th rep-embedding, at layer $l$, the rep-points are derived the Cartesian product $\prod_{dim=1}^3 (\bm{\mu}^{l,k,dim}_s-3\bm{\sigma}^{l,k,dim}_s, \cdots, \bm{\mu}^{l,k,dim}_s+3\bm{\sigma}^{l,k,dim}_s)$. In total, there are $7^3=343$ rep-points sampled per mean. Comparisons of performance and inference speed are provided in Section~\ref{sec:experiments}.

\subsubsection{Group Cross-Attention}
The group cross-attention (GCA) captures the visual features that correspond to each subject and object rep-embedding by computing their attention scores, respectively. The application of GCA involves performing separate interactions between queries and subject rep-point features, as well as between keys and object rep-point features. The subject GCA for the $i$-th entity is defined as
\begin{equation}
\begin{aligned}
   \rmQ^{l,i} & = \text{GCA}(\vq, \vk, \vv) \\
     & = \text{softmax}( \vq \vk^{T} / \sqrt{d_G}) \; \vv \\
   \vq & = \text{Linear}(\rmQ^{l-1,i} + \rmQ_{b}^i) \in \R^{h_{G} \times K \times d_{G}} \\
   \vk & = \text{Linear}(\rmV^{l,i}_s + \rmP\rmE^{l,i}_s) \in \R^{h_G \times K \times m \times d_G} \\
   \vv & = \text{Linear}(\rmV^{l,i}_s) \in \R^{h_G \times K \times m \times d_G},
 \end{aligned}
\label{eq:gca}
\end{equation}
where $i$ indexes the entity, $\text{Linear}(\cdot)$ is a fully-connected layer ($\vq$, $\vk$, and $\vv$ are projected with different parameters), $h_G$ is the number of attention heads, and $d_G$ is the dimension of each head. 
GCA is performed independently among groups in parallel, where the cross-attention between a rep-embedding and its corresponding sampled rep-point features are performed. 
Following the transformer architecture~\cite{vaswani2017attention}, multi-head outputs are concatenated and projected with a fully-connected layer, and a residual connection~\cite{he2016deep} with layer normalization~\cite{ba2016layer} is added. 
Likewise, another GCA layer with different parameters is performed for keys as $\rmK^l = \text{GCA} \left(\text{Linear}(\rmK^{l-1} + \rmK_{b}), \text{Linear}(\rmV^{l}_o + \rmP\rmE^{l}_o), \text{Linear}(\rmV^{l}_o) \right)$.
GCA allows each rep-embedding to focus on different visual features, carrying the relevant ones along the way.

\subsubsection{Two-Way Relational Cross-Attention} \label{sec:rca}

In GCA, queries and keys are updated by their corresponding sampled features. In a two-way relational cross-attention (RCA) layer, queries are updated by keys, and vice versa. Firstly, the raw attention weights $\rmA^l$ are computed between flattened $\rmQ^l$ and $\rmK^l$. Since the projections of queries and keys are different, the attention weights are not symmetric and can be normalized along different dimensions. Softmax is then applied on $\rmA^l$ along the dimension of keys, and along the dimension of queries to obtain two-way attention weights. Finally, queries and keys are updated by multiplying the corresponding attention weights with values. The two-way relational cross-attention is formally defined as following:
\begin{equation}
\begin{aligned}
   \rmQ^{l}, \rmK^l & = \text{RCA}(\vq, \vk, \vv_q, \vv_k) \\
   \vq & = \text{Linear}(\rmQ^{l} + \rmQ_{b}) \\
   \vk & = \text{Linear}(\rmK^{l} + \rmK_{b}) \\
   \vv_q & = \text{Linear}(\rmQ^{l}) \\
   \vv_k & = \text{Linear}(\rmK^{l}) \\
   \rmA^l & = \vq \vk^{T} / \sqrt{d_R} \in \R^{h_R \times n_q \times n_k} \\
   \rmA^l_k &  = \text{softmax}(\rmA^l) \; s.t. \; \sum\nolimits_{j=1}^{n_k} \rmA^{l, *, *, j}_k=1 \\
   \rmA^l_q &  = \text{softmax}(\rmA^l) \; s.t. \; \sum\nolimits_{i=1}^{n_q} \rmA^{l, *, i, *}_q=1 \\
   \rmQ^{l} & = \rmA^l_k \vv_k, \; \rmK^{l} = (\rmA^l_q)^T \vv_q,
 \end{aligned}
\label{eq:rca}
\end{equation}
where $n_q = n_k = n \times K$, $h_R$ is the number of attention heads, and $d_R$ is the dimension of each head, ``$*$'' denotes any index along the specific dimension. Additionally, two MLPs are used for projecting the output queries and keys respectively, following the feed-forward network design in \cite{vaswani2017attention}. Without abuse of notation, the notations of output queries $\rmQ^l$ and keys $\rmK^l$ of RCA layers remain the same. After $L$ decoder layers, the outputs $\rmQ^L$ and $\rmK^L$ are obtained which are used for predicate prediction. 

\subsection{Relationships as Attention Weights}

For most SGG methods using either box-based, or query-based representation, predicate classification is performed on triplet features in different forms of feature fusions. For example, box-based methods use the concatenation of pairwise entity features and their union-box features, and query-based methods use learnable triplet embeddings to perform classification directly. Neither of them can capture the directional information of scene graphs explicitly which could cause learning bias and overfitting on dominant visual configurations. A very simple case is that there is a common triplet \texttt{<man, on, street>} in the dataset, the learnt model will likely predict \texttt{on} if it detects \texttt{man} and \texttt{street} concurrently. Conversely, triplets such as \texttt{<man, standing on, street>} and \texttt{<street, under, man>} are considered as incorrect predictions and are treated as negative examples during training, despite being semantically valid triplets. The presence of rare, bidirectional~\cite{abou2022topology}, and unannotated relationships~\cite{zhang2022fine} hinders the learning of representative semantics for SGG methods.

Similar to the RCA discussed in Section~\ref{sec:rca}, $\rmQ^L$ and $\rmK^L$ are projected with $h_A$ heads, and each head has $d_A$ dimensions. Unnormalized attention weights $\rmA^L \in \R^{h_A \times n_q \times n_k}$ are computed between projected queries and keys. Different from RCA, the multi-head attention weights are not multiplied by projected values. The asymmetric nature of dot-product attention serves as a natural metric for quantifying the relationship between subjects and objects. Moreover, attention weights of each head captures distinct semantics, similar to the way feature channels operate. Therefore, the attention weights can be mapped to the predicate classification $ \rmY \in \R^{P \times n_q \times n_k}$ via a fully-connected layer, where $P$ is the number of predicate classes of a dataset. In parallel, a binary relation mask $\rmH \in \R^{n_q \times n_k}$ is also predicted to classify if a relationship exists between a pair of rep-embeddings. The relation mask is used for suppressing low-quality predictions. Formally, $\rmY$ and $\rmH$ are obtained as
\begin{equation}
\begin{aligned}
    \rmY & = \text{Linear}(\rmA^L) \\
   \rmH & = \text{Linear}(\rmA^L) \\
   \rmA^{L} & = \vq \vk^T / \sqrt{d_A} + \vb_A \\
   \vq & = \text{Linear}(\rmQ^{L} + \rmQ_{b}) \\
   \vk & = \text{Linear}(\rmK^{L} + \rmK_{b}),
 \end{aligned}
\end{equation}
where $\vb_A$ is an added bias term. $\rmY$ represents relationships between subject and object rep-embeddings instead of subjects and objects. To get pairwise relationships between entities, $\rmY$ is re-arranged to $ \R^{P \times n \times n \times K^2} $. During training, Gumbel-Softmax~\cite{DBLP:conf/iclr/JangGP17} is applied over the last dimension of $\rmY$ to sample the rep-embedding pairs with the largest logits, and $\rmY$ is reduced to the shape of $ \R^{P \times n \times n} $. An annealing schedule is applied that changes the Gumbel-Softmax temperature from 10 to 0.5 gradually through the first 30\% iterations. During inference, the maximum logits over the last dimension are chosen. The relation mask $\rmH$ undergoes the same re-arrangement operation, followed by selecting the maximum value for both training and inference. The final predicate classification score is defined as $\sqrt{ \sigma(\rmH) \cdot \sigma(\rmY) }$, where $\sigma(\cdot)$ is the sigmoid function.

\subsection{Training}

In this section, we discuss the training losses and strategies to address the challenges posed by long-tailed sparsely-annotated data. The hyper-parameters and losses for entity detection remain the same as in FCOS~\cite{tian2020fcos}.

\subsubsection{Losses} \label{sec:losses}

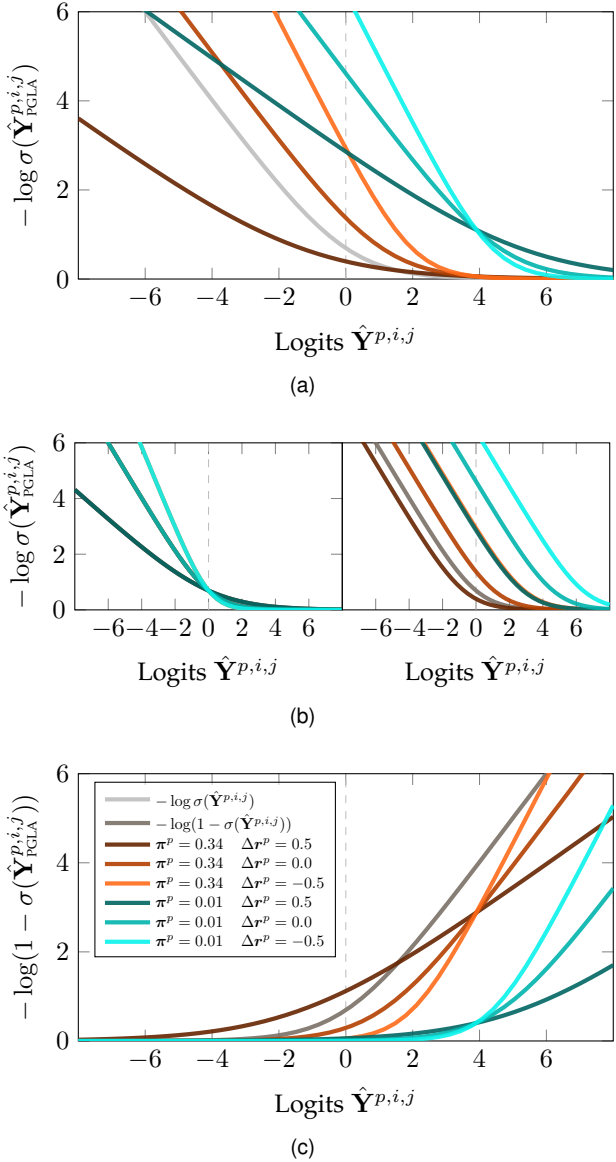
\begin{figure}[t]
\centering
\subfloat[]{%
\tikzmath{
\uniform = ln(0.02);
\d1 = 0.0;
\d2 = -1.0;
\d3 = 1.0;
\r1 = 0.5;
\r2 = 0.0;
\f1 = 0.34;
\f2 = 0.01;
\f3 = 0.01;
}
\pgfplotsset{width=\columnwidth,compat=1.18}
\begin{tikzpicture}[
declare function={
bce(\x)=-ln(1/(1+exp(-\x))+1e-8);
negbce(\x)=-ln(1-1/(1+exp(-\x))+1e-8);
sigmoid(\x)=1/(1+exp(-\x));
weights(\f,\r)=tanh(-\r)+1;
diff(\d,\r) = (1-\r^2)*(\d * exp(\d));
bias(\f,\r,\d) = ln(\f) + tanh(-\r) * \uniform + \d;
}]
\begin{axis}[
scale only axis,
width=0.8\columnwidth,
height=0.4\columnwidth,
xmin=-8, xmax=8, ymin=0, ymax=6, samples=50,
domain=-8:8,
legend style={fill=white, fill opacity=0.6, draw opacity=1, text opacity=1,nodes={scale=0.6, transform shape}},
legend columns=1, 
legend cell align={left},
xlabel={Logits $\hat{\rmY}^{p,i,j}$},
ylabel={$ - \log \sigma(\hat{\rmY}^{p,i,j}_{\text{\tiny{PGLA}}}) $}, 
legend pos=north east,
xtick={-6,-4,...,6},
ytick={0,2,...,8},
extra x ticks={0},
extra x tick style={
    grid=major,
    grid style=dashed,
    ticklabel style={opacity=0},
    tick style={draw=none},
    xticklabel=\empty
},
extra y tick style={
    ticklabel style={opacity=0},
    tick style={draw=none},
    xticklabel=\empty
},
]
  \addplot[opacity=0.9, color={rgb,255:red,190;green,190;blue,190}, ultra thick] {bce(x)};
  \addplot[opacity=0.9, color={rgb,255:red,102;green,36;blue,0}, ultra thick]{bce(weights(\f1,\r1) * x + bias(\f1,\r1,\d1))};
  \addplot[opacity=0.9, color={rgb,255:red,179;green,63;blue,0}, ultra thick] {bce(weights(\f1,\r2) * x + bias(\f1,\r2,\d1))};
  \addplot[opacity=0.9, color={rgb,255:red,255;green,107;blue,26}, ultra thick] {bce(weights(\f1,-\r1) * x + bias(\f1,-\r1,\d1))};

  \addplot[opacity=0.9, color={rgb,255:red,0;green,102;blue,99}, ultra thick] {bce(weights(\f2,\r1) * x + bias(\f2,\r1,\d1))};
  \addplot[opacity=0.9, color={rgb,255:red,0;green,179;blue,173}, ultra thick] {bce(weights(\f2,\r2) * x + bias(\f2,\r2,\d1))};
  \addplot[opacity=0.9, color={rgb,255:red,10;green,242;blue,235}, ultra thick] {bce(weights(\f2,-\r1) * x + bias(\f2,-\r1,\d1))};
  
\end{axis}
\end{tikzpicture}%
}

\subfloat[]{%
\tikzmath{
\uniform = ln(0.02);
\d1 = 0.0;
\d2 = -1.0;
\d3 = 1.0;
\r1 = 0.5;
\r2 = 0.0;
\f1 = 0.34;
\f2 = 0.01;
\f3 = 0.01;
}

\begin{tikzpicture}[
baseline,
trim axis right,
declare function={
bce(\x)=-ln(1/(1+exp(-\x))+1e-8);
negbce(\x)=-ln(1-1/(1+exp(-\x))+1e-8);
sigmoid(\x)=1/(1+exp(-\x));
weights(\f,\r)=tanh(-\r)+1;
diff(\d,\r) = (1-\r^2)*(\d * exp(\d));
bias(\f,\r,\d) = 0;
}]
\begin{axis}[
width=0.4\columnwidth,
height=0.25\columnwidth,
scale only axis,
xmin=-8, xmax=8, ymin=0, ymax=6, samples=50,
domain=-8:8,
legend style={fill=white, fill opacity=0.6, draw opacity=1, text opacity=1,nodes={scale=0.6, transform shape}},
legend columns=1, 
legend cell align={left},
xlabel={Logits $\hat{\rmY}^{p,i,j}$},
ylabel={$ - \log \sigma(\hat{\rmY}^{p,i,j}_{\text{\tiny{PGLA}}}) $}, 
legend pos=north east,
xtick={-6,-4,...,6},
ytick={0,2,...,8},
extra x ticks={0},
extra x tick style={
    grid=major,
    grid style=dashed,
    ticklabel style={opacity=0},
    tick style={draw=none},
    xticklabel=\empty
},
extra y tick style={
    ticklabel style={opacity=0},
    tick style={draw=none},
    xticklabel=\empty
},
]
  \addplot[opacity=0.9, color={rgb,255:red,122;green,112;blue,103}, ultra thick] {bce(x)};

  \addplot[opacity=0.9, color={rgb,255:red,102;green,36;blue,0}, ultra thick]{bce(weights(\f1,\r1) * x + bias(\f1,\r1,\d1))};

  \addplot[opacity=0.9, color={rgb,255:red,179;green,63;blue,0}, ultra thick] {bce(weights(\f1,\r2) * x + bias(\f1,\r2,\d1))};
  \addplot[opacity=0.9, color={rgb,255:red,255;green,107;blue,26}, ultra thick] {bce(weights(\f1,-\r1) * x + bias(\f1,-\r1,\d1))};

  \addplot[opacity=0.9, color={rgb,255:red,0;green,102;blue,99}, ultra thick] {bce(weights(\f2,\r1) * x + bias(\f2,\r1,\d1))};

  \addplot[opacity=0.9, color={rgb,255:red,0;green,179;blue,173}, ultra thick] {bce(weights(\f2,\r2) * x + bias(\f2,\r2,\d1))};  
  \addplot[opacity=0.9, color={rgb,255:red,10;green,242;blue,235}, ultra thick] {bce(weights(\f2,-\r1) * x + bias(\f2,-\r1,\d1))};

\end{axis}
\end{tikzpicture}%
\begin{tikzpicture}[
baseline,
trim axis left,
declare function={
bce(\x)=-ln(1/(1+exp(-\x))+1e-8);
negbce(\x)=-ln(1-1/(1+exp(-\x))+1e-8);
sigmoid(\x)=1/(1+exp(-\x));
weights(\f,\r)=1;
diff(\d,\r) = (1-\r^2)*(\d * exp(\d));
bias(\f,\r,\d) = ln(\f) + tanh(-\r) * \uniform + \d;
}]
\begin{axis}[
width=0.4\columnwidth,
height=0.25\columnwidth,
scale only axis,
xmin=-8, xmax=8, ymin=0, ymax=6, samples=50,
domain=-8:8,
legend style={fill=white, fill opacity=0.6, draw opacity=1, text opacity=1,nodes={scale=0.6, transform shape}},
legend columns=1, 
legend cell align={left},
xlabel={Logits $\hat{\rmY}^{p,i,j}$},
legend pos=north east,
xtick={-6,-4,...,6},
ymajorticks=false,
extra x ticks={0},
extra x tick style={
    grid=major,
    grid style=dashed,
    ticklabel style={opacity=0},
    tick style={draw=none},
    xticklabel=\empty
},
extra y tick style={
    ticklabel style={opacity=0},
    tick style={draw=none},
    xticklabel=\empty
},
]
  \addplot[opacity=0.9, color={rgb,255:red,122;green,112;blue,103}, ultra thick] {bce(x)};

  \addplot[opacity=0.9, color={rgb,255:red,102;green,36;blue,0}, ultra thick]{bce(weights(\f1,\r1) * x + bias(\f1,\r1,\d1))};

  \addplot[opacity=0.9, color={rgb,255:red,179;green,63;blue,0}, ultra thick] {bce(weights(\f1,\r2) * x + bias(\f1,\r2,\d1))};

  \addplot[opacity=0.9, color={rgb,255:red,255;green,107;blue,26}, ultra thick] {bce(weights(\f1,-\r1) * x + bias(\f1,-\r1,\d1))};

  \addplot[opacity=0.9, color={rgb,255:red,0;green,102;blue,99}, ultra thick] {bce(weights(\f2,\r1) * x + bias(\f2,\r1,\d1))};

  \addplot[opacity=0.9, color={rgb,255:red,0;green,179;blue,173}, ultra thick] {bce(weights(\f2,\r2) * x + bias(\f2,\r2,\d1))};

  \addplot[opacity=0.9, color={rgb,255:red,10;green,242;blue,235}, ultra thick] {bce(weights(\f2,-\r1) * x + bias(\f2,-\r1,\d1))};

\end{axis}
\end{tikzpicture}%
}

\subfloat[]{%
\tikzmath{
\uniform = ln(0.02);
\d1 = 0.0;
\d2 = -1.0;
\d3 = 1.0;
\r1 = 0.5;
\r2 = 0.0;
\f1 = 0.34;
\f2 = 0.01;
\f3 = 0.01;
}
\pgfplotsset{width=\columnwidth,compat=1.18}
\begin{tikzpicture}[
declare function={
bce(\x)=-ln(1/(1+exp(-\x))+1e-8);
negbce(\x)=-ln(1-1/(1+exp(-\x))+1e-8);
sigmoid(\x)=1/(1+exp(-\x));
weights(\f,\r)=tanh(-\r)+1;
diff(\d,\r) = (1-\r^2)*(\d * exp(\d));
bias(\f,\r,\d) = ln(\f) + tanh(-\r) * \uniform + \d;
}]
\begin{axis}[
scale only axis,
width=0.8\columnwidth,
height=0.4\columnwidth,
xmin=-8, xmax=8, ymin=0, ymax=6, samples=50,
domain=-8:8,
legend style={fill=white, fill opacity=0.6, draw opacity=1, text opacity=1,nodes={scale=0.6, transform shape}},
legend columns=1, 
legend cell align={left},
xlabel={Logits $\hat{\rmY}^{p,i,j}$},
ylabel={$ - \log (1 - \sigma(\hat{\rmY}^{p,i,j}_{\text{\tiny{PGLA}}})) $}, 
legend pos=north west,
xtick={-6,-4,...,6},
ytick={0,2,...,8},
extra x ticks={0},
extra x tick style={
    grid=major,
    grid style=dashed,
    ticklabel style={opacity=0},
    tick style={draw=none},
    xticklabel=\empty
},
extra y tick style={
    ticklabel style={opacity=0},
    tick style={draw=none},
    xticklabel=\empty
},
]
    \addlegendimage{opacity=0.9, ultra thick, color={rgb,255:red,190;green,190;blue,190}}
    \addlegendentry{$- \log \sigma(\hat{\rmY}^{p,i,j})$}
  \addplot[opacity=0.9, color={rgb,255:red,122;green,112;blue,103}, ultra thick] {negbce(x)};
  \addlegendentry{$ - \log (1 - \sigma(\hat{\rmY}^{p,i,j})) $}
  \addplot[opacity=0.9, color={rgb,255:red,102;green,36;blue,0}, ultra thick]{negbce(weights(\f1,\r1) * x + bias(\f1,\r1,\d1))};
  \addlegendentry{\( \bm\pi^p=\f1 \;\;\;\; \Delta \vr^p=\r1\)}
  \addplot[opacity=0.9, color={rgb,255:red,179;green,63;blue,0}, ultra thick] {negbce(weights(\f1,\r2) * x + bias(\f1,\r2,\d1))};
  \addlegendentry{\( \bm\pi^p=\f1 \;\;\;\; \Delta \vr^p=\r2 \)}
  \addplot[opacity=0.9, color={rgb,255:red,255;green,107;blue,26}, ultra thick] {negbce(weights(\f1,-\r1) * x + bias(\f1,-\r1,\d1))};
  \addlegendentry{\( \bm\pi^p=\f1 \;\;\;\; \Delta \vr^p=-\r1 \)}

  \addplot[opacity=0.9, color={rgb,255:red,0;green,102;blue,99}, ultra thick] {negbce(weights(\f2,\r1) * x + bias(\f2,\r1,\d1))};
  \addlegendentry{\( \bm\pi^p=\f2 \;\;\;\; \Delta \vr^p=\r1 \)}
  \addplot[opacity=0.9, color={rgb,255:red,0;green,179;blue,173}, ultra thick] {negbce(weights(\f2,\r2) * x + bias(\f2,\r2,\d1))};
  \addlegendentry{\( \bm\pi^p=\f2 \;\;\;\; \Delta \vr^p=\r2 \)}
  \addplot[opacity=0.9, color={rgb,255:red,10;green,242;blue,235}, ultra thick] {negbce(weights(\f2,-\r1) * x + bias(\f2,-\r1,\d1))};
  \addlegendentry{\( \bm\pi^p=\f2 \;\;\;\; \Delta \vr^p=-\r1 \)}
  
\end{axis}
\end{tikzpicture}%
}
\caption{BCE loss on PGLA-adjusted logits. (a) Loss on a positive predicate $p$ where $\rmY^{p,i,j} = 1$. (b) Loss on a positive predicate $p$ with $\rmB = \bm{0}$ (left), and with $\rmW = \bm{1}$ (right). Legend follows (c) Loss on a negative predicate $p$ where $\rmY^{p,i,j} = 0$. The legend is shared across (a) - (c), and gray lines in figures represent the BCE loss on original logits.}
\label{fig:la}
\end{figure}

Due to the potential existence of multiple relationships (directional or bidirectional) between two entities, the scene graph frequently exhibits a multi-graph structure. The predicate classification is considered as a multi-label multi-class classification problem. Consequently, we use the binary cross-entropy (BCE) instead of softmax to supervise the predicate classification and relation mask. To balance well-learned and hard examples, the focal loss (FL)~\cite{lin2017focal} for BCE is used. Specifically, given the predicted logits $\hat{\rmY}$ and ground-truth labels $\rmY \in \{ 0,1 \}^{P \times n \times n}$, the focal BCE is defined as
\begin{equation} \label{eq:fl}
\resizebox{0.95\hsize}{!}{$%
\text{FL}(\hat{\rmY}, \rmY) =
\begin{cases}
 -\frac{\alpha}{N_{pos}} \sum\limits_{p,i,j} (1-\sigma(\hat{\rmY}^{p,i,j}))^\gamma\log(\sigma(\hat{\rmY}^{p,i,j})), & \rmY^{p,i,j}=1\\
-\frac{1-\alpha}{N_{pos}} \sum\limits_{p,i,j} \sigma(\hat{\rmY}^{p,i,j})^\gamma\log(1-\sigma(\hat{\rmY}^{p,i,j})), &  \rmY^{p,i,j}=0,
\end{cases}
$%
}%
\end{equation}
where $\alpha$ is a class-balance weighting factor, $\gamma$ is the focal factor, $p$ indexes the predicate classes, $i$ indexes subjects, $j$ indexes objects, and $N_{pos}=\sum_{p,i,j} \rmY^{p,i,j}$ is the number of ground-truth triplets. As the predicate prediction forms a fully-connected graph among $n$ entities, the ground-truth $\rmY$ is inherently sparse with few ones. 
We select $\alpha=0.75$ to prioritize generating larger logits for positive predicates, rather than penalizing negative predicates. For the focal factor $\gamma$, we set it differently based on the predicate frequency of training data. Let the predicate priors be $\bm{\pi}$, \eg, the empirical predicate class frequencies in the training dataset, we compute a predicate-specific $\bm{\gamma}$ to replace $\gamma$ in (\ref{eq:fl}) as
\begin{equation}
\bm\gamma = \gamma \cdot \frac{\bm{\pi} - \min(\bm\pi)}{\max(\bm\pi) - \min(\bm\pi)},
\end{equation}
where $\gamma = 2$, $\min(\cdot)$ and $\max(\cdot)$ are operations to get the minimum and maximum value respectively. For the tail predicates, $\bm{\gamma}^p$ is small so that it encourages the logits to be larger. For the head predicates, $\bm{\gamma}^p$ becomes larger and down-weights the loss. For supervising the relation mask, we empirically select $\alpha=0.75$ and $\gamma=2$.

As discussed in \cite{newell2017pixels,liu2021fully,zhang2022fine}, the sparsity of data annotations for relations, coupled with the presence of numerous unannotated ones, leads to semantic ambiguity and poses challenges during training. Therefore, simply considering triplets without ground-truth annotations as negative is not optimal. For training the relation mask $\hat{\rmH} \in \R^{n \times n}$ among $n$ entities, we sub-sample the negative triplets with a ratio of 10:1 in proportion to the number of ground-truth triplets. Additional, we employ a margin ranking loss for predicted relation classification $\hat{\rmY}$. Instead of supervising negative samples (where $\rmY^{p,i,j} = 0, \forall p $) with labels of zero in BCE, per-predicate margins are calculated and used as the upper bounds for negative samples' logits. The margin ranking loss $ \mathcal{L}_{\eta} $ is defined as
\begin{equation}
\begin{aligned}
 \mathcal{L}_{\eta}(\hat{\rmY}, \rmY) & =  \frac{1}{N_{neg}} \sum\limits_{p,i,j} \max(\sigma(\hat{\rmY}^{p,i,j}) - \eta^p), \; 0), \\
 \eta^p & = \min \{ \sigma (\hat{\rmY}^{p,i,j}) \;|\; \rmY^{p,i,j} = 1 ,\; \forall \; i, j\},
\end{aligned}
\end{equation}
where $N_{neg}$ is the number of negative samples, and $\eta^p$ is the margin for predicate $p$ between positive and negative samples. To ensure a normalized effect across different scenes (images), the margins are computed on a per-image basis. By employing this loss, unannotated triplets are neither excessively penalized, which could lead to training ambiguity, nor overly encouraged, which could result in lower ranks for positive triplets.

For sampling useful subject and object rep-points, margin ranking losses are applied to the subject and object offset means $\bm{\mu}^l_s$ and $\bm{\mu}^l_o$, for $l=1,...,L$.
For the rep-point coordinates reparameterized by the mean, their margins are the top-left and bottom-right corner coordinates of union bounding boxes of triplets involving the entity.
Consequently, the loss is nonzero when a mean rep-point is outside of a corresponding union bounding box.

\subsubsection{Performance-Guided Logit Adjustment}

Building upon the concept of the logit adjustment (LA)~\cite{menon2021longtail}, we introduce the run-time performance-guided logit adjustment (PGLA) as a novel approach to enhance the performance on long-tail problems. In logit adjusted softmax cross-entropy~\cite{menon2021longtail}, the logits from a classifier are added with a bias term $\log \bm\pi$ to create pairwise margins between classes. Due to the intricacies involved in entity detection and scene graph generation, employing a fixed bias throughout the training process is suboptimal. Hence, we extend the logit adjustment to a more general form of affine transformation by adding a weight term, and leveraging the training statistics to more accurately quantify the class margins.

In the context of the scene graph generation task, recall is used to guide the strength of adjustment, and PGLA is only applied to logits of positive relations. For the pair of subject $i$ and object $j$, the predicted logits are adjusted as
\begin{equation} \label{eq:la}
\hat{\rmY}^{:,i,j}_{\text{\tiny{PGLA}}} = 
\begin{cases}
\rmW \odot \hat{\rmY}^{:,i,j} + \rmB & (\exists \; p)\; \rmY^{p,i,j} = 1\\
\hat{\rmY}^{:,i,j} & \text{otherwise,}
\end{cases}
\end{equation}
where $\rmW \in \R^{P}$ and $\rmB \in \R^{P}$ are the weight and bias factors, respectively, and the operation denoted by ``$:$'' selects all elements along the specified dimension. By setting $\rmW = \bm{1}$ and $\rmB = \log \bm\pi$, (\ref{eq:la}) yields logit adjustment~\cite{menon2021longtail}. The weight and bias factors for (\ref{eq:la}) are calculated as
\begin{equation} \label{eq:w_b}
\begin{aligned}
   \rmW & = -  \tanh ( \Delta \vr) + 1, \\
   \rmB & =  - \tanh (\Delta \vr / \lambda) \cdot \log (P^{-1}) + \log \bm\pi,
 \end{aligned}
\end{equation}
where $\Delta \vr = \vr - \bar{\vr}$, $\vr$ is the measured recall, $\bar{\vr}$ is the mean of the recall, $\lambda$ is a hyper-parameter set to 1 by default, and $\log (P^{-1})$ is the log probability of the uniform distribution over $P$ predicates. The hyper-parameter $\lambda$ controls the sensitivity of PGLA \wrt the recall differences. A smaller value of $\lambda$ increases the sensitivity, and more losses are enforced for predicates with low recall. The tanh($\cdot$) function limits $\Delta \vr$ or $\Delta \vr / \lambda$ within the range of [-1, 1], and there exist alternative functions that serve the same purpose.

The effect of the long-tailed problem on losses can be described as follows: tail predicates play the role of negative classes when training head predicates, and constantly receive losses for being classified as negatives during training; on the other hand, head predicates receive more losses for being positive and less losses for being classified as negatives during training.
To address this effect, there are various considerations to be taken into account regarding $\rmW$ and $\rmB$. 
Overall, the impacts of $\rmW$ and $\rmB$ \wrt the BCE loss with adjusted logits $\hat{\rmY}_{\text{\tiny{PGLA}}}$ are illustrated in Fig.~\ref{fig:la}.  
In the case where a predicate $p$ achieves a relatively higher recall ($\Delta \vr^p > 0 $), we decrease the value of $\rmW^p$ to encourage the network to generate larger logits when $p$ is positive ($\rmY^{p,*,*} = 1$), and smaller logits when $p$ is negative ($\rmY^{p,*,*} = 0$). This adjustment of $\rmW$ tries to push the predictions towards the saturation regions of the sigmoid function so that it is easier to distinguish between positive and negative classes. Simultaneously, as the recall $\vr^p$ increases relatively, $\rmB^p$ increases and less loss is assigned to predicate $p$, allowing us to focus on other predicates with lower recall. In addition to utilizing the recall, the bias factor per predicate will be adjusted based on its prior distribution as well. A tail predicate is assigned with a smaller bias factor, and receive larger losses when being positive and smaller loss when being negative. 

Despite achieving the goal of assigning class margins dynamically to address the long-tailed problem, it is critical to note that the similarities between predicates have not been taken into account. Therefore, we introduce a training statistic named ``confusion logits'', denoted as $\rmD \in \R^{ P \times P}$, which tracks pairwise predicate logit differences if the predictions are incorrect. The confusion logit between the ground-truth predicate $p$ and an arbitrary predicate $\hat{p}$ is computed as
\begin{equation} \label{eq:confusion_logits}
\begin{aligned}
 \rmD^{p, \hat{p}} = & \underset{(p,i,j) \in \Omega}{\mathrm{mean}} \{ \vd_{logits}^{i,j} \cdot  \vd_{\pi}^{p,\hat{p}} \} \\
& \Omega  \coloneqq \{ (p,i,j) \;|\;  \rmY^{p,i,j} = 1 \} \\
 \vd_{logits}^{i,j} = &  \mathrm{ReLU} ( \hat{\rmY}^{\hat{p},i,j} - \hat{\rmY}^{p,i,j} ) \\
 \vd_{\pi}^{p,\hat{p}} = & \tanh(\mathrm{ReLU}(\log \bm\pi^{\hat{p}} - \log \bm\pi^p)),
\end{aligned}
\end{equation}
where $\mathrm{ReLU}(\cdot)$~\cite{nair2010rectified} is used for selecting positive values only.
The confusion logits for a GT predicate $p$ are computed only with respect to incorrectly-predicted predicates ($ \hat{\rmY}^{\hat{p},i,j} > \hat{\rmY}^{p,i,j} $) with larger priors ($\vd_{\pi}^{p,\hat{p}} > 0$ if $\bm\pi^{\hat{p}} > \bm\pi^p$). Large value of $\rmD^{p, \hat{p}}$ means that $p$ is often mis-classified as $\hat{p}$. The confusion logits not only quantify the effects of long-tailed data, but also the semantic similarities between predicates. For similar tail predicates like \texttt{across} and \texttt{along}, their confusion logits can be large as well.
During training, per-instance PGLA is applied instead by utilizing the corresponding confusion logits of a specific ground-truth predicate $p$, and (\ref{eq:la}) is modified as
\begin{equation} \label{eq:la2}
\hat{\rmY}^{:,i,j}_{\text{\tiny{PGLA}}} = 
\begin{cases}
\rmW \odot \hat{\rmY}^{:,i,j} + \rmB + \rmD^{p,:} & (\forall \; p)\; \rmY^{p,i,j} = 1\\
\hat{\rmY}^{:,i,j} & \text{otherwise.}
\end{cases}
\end{equation}

Ideally, recall of each predicate should be close to the mean recall for a balanced classifier. 
However, head classes tend to exhibit significantly higher recall compared to tail classes due to long-tailed training data. Therefore, we evaluate the recall per predicate differently. 
The exponential moving average (EMA) of predicate recall per mini-batch is computed to estimate the performance change over time. 
For each image in the mini-batch, the histogram of ground-truth relationships per predicate is obtained as $\vn$. 
Next, based on the top $ \mathrm{sum} (\vn) $ triplets and predicate priors, ranking targets $\bm\kappa$ are set differently per predicate. 
For a predicate $p$, the top-$\bm\kappa^p$ triplets is used for computing the recall, where tail predicates are assigned with smaller $\bm\kappa^p$ while head predicates are assigned with larger values.
By setting distinct ranking targets for each predicate, it forces tail predicates to rank higher than head ones.
Finally, the per-predicate recall $\vr$ is calculated. A EMA momentum specific to each predicate is assigned as $\bm\rho=0.9999^{-\log\bm\pi}$. 
Assume the batch size is 1, the process of calculating recall at iteration $t$ is detailed in Algorithm~\ref{alg:recall}. 
The confusion logits $\rmD_t$ is calculated and updated per iteration via EMA as well. 
Consequently, the PGLA can be performed given the run-time values of $\rmW_t$, $\rmB_t$, and $\rmD_t$ in (\ref{eq:la2}). 

\begin{algorithm}[t]
  \caption{Recall Calculation at training iteration $t$}
  \label{alg:recall}
  \begin{algorithmic}[1]
    \Require{$\hat{\rmY}$, $\rmY$, $\vr_{t-1}$, $\bm\pi$}
    \Ensure{$\vr_t$}
    \Let{$\vr_{t}$}{$\bm{0}_P$}
    \Let{$\bm\nu$}{$ \arg \mathrm{sort}(\bm\pi)$} \Comment{indices of sorted predicate}
    \Let{$\vn$}{$\sum_{i,j} \rmY^{:, i, j}$} \Comment{No. of GTs per predicate}
    \Let{$\bm\kappa$}{$\mathrm{cumsum}(\vn[\bm\nu])$} \Comment{cumulative sum of No. of GTs}
    
    \For{$p \gets \bm\nu^1 \text{ to } \bm\nu^P$}
        \Let{$\hat{R}_{\mathrm{top-}\bm\kappa^p}$}{$\arg\max_{\bm\kappa^p}$($ \hat{\rmY}$)} \Comment{top $\bm\kappa^p$ triplets}
        \Let{$R^p$}{$ \mathrm{nonzero} ( \rmY^p )$} \Comment{GT triplets}
        \Let{$\vr^p_{t}$}{$\vr^p_{t} + \mathrm{match} ( \hat{R}_{\mathrm{top-}\bm\kappa^p}$, $R^p ) $} \Comment{accumulate matches}
      \EndFor
      \Let{$\vr_{t}$}{$\vr_{t} \oslash \vn$} \Comment{element-wise division}
      \Let{$\vr_{t}$}{$ (1 - \bm\rho) \cdot \vr_{t} + \bm\rho \cdot \vr_{t-1}$} \Comment{EMA}
      \State \Return{$\vr_{t}$}
  \end{algorithmic}
\end{algorithm}




\section{Experiments}\label{sec:experiments}

\begin{table*}[t]
\centering
\resizebox{\linewidth}{!}{%
\begin{threeparttable}
\caption{Comparisons of R@K and mR@K results on VG150 between the proposed methods and SOTA methods. Methods are grouped from top to bottom as: point-based, query-based, and box-based methods. FCSGG~\cite{liu2021fully} uses HRNet~\cite{wang2020deep} as backbone, and CoRF~\cite{Adaimi_2023_WACV} uses Swin-S~\cite{liu2021swin}. The best results are bold, and the second-best results are underlined.}
\label{tab:sota}

\sisetup{table-format=2.1, table-number-alignment=center, table-space-text-post=\hspace{0pt}}
\setlength{\tabcolsep}{4pt}
\renewcommand{\arraystretch}{1.1}
    \begin{tabular}{l | 
    SSS c@{\hspace{5pt}} SSS | 
    SSS c@{\hspace{5pt}} SSS | 
    SSS c@{\hspace{5pt}} SSS } 
    \toprule 
      & \multicolumn{7}{c|} {Predicate Classification} & \multicolumn{7}{c|} { Scene Graph Classification } & \multicolumn{7}{c} { Scene Graph Detection } \\ 
    & \multicolumn{3}{c}{R@20/50/100} & & \multicolumn{3}{c|}{mR@20/50/100} & \multicolumn{3}{c}{R@20/50/100} & & \multicolumn{3}{c|}{mR@20/50/100} & \multicolumn{3}{c}{R@20/50/100} & & \multicolumn{3}{c}{mR@20/50/100} \\
    \midrule

    FCSGG~\cite{liu2021fully} & 33.4 & 41.0 & 45.0 & & 4.9 & 6.3 & 7.1 & 19.0 & 23.5 & 25.7 & & 2.9 & 3.7 & 4.1 & 16.1 & 21.3 & 25.1 & & 2.7 & 3.6 & 4.2 \\

    CoRF~\cite{Adaimi_2023_WACV} & \multicolumn{1}{c}{-} & 45.4 & \multicolumn{1}{c}{-} & & \multicolumn{1}{c}{-} & 10.1 & \multicolumn{1}{c|}{-} & \multicolumn{1}{c}{-} & 18.7 & \multicolumn{1}{c}{-} & & \multicolumn{1}{c}{-} & 3.9 & \multicolumn{1}{c|}{-} & \multicolumn{1}{c}{-} & 18.6 & \multicolumn{1}{c}{-} & & \multicolumn{1}{c}{-} & 3.9 & \multicolumn{1}{c}{-} \\ 
    
    \midrule
    
    RelTR~\cite{cong2023reltr} \tnote{\textdaggerdbl} & \textbf{63.1} & \underline{64.2} & \multicolumn{1}{c}{-} & & 20.0 & 21.2 & \multicolumn{1}{c|}{-} & 29.0 & 36.6 & \multicolumn{1}{c}{-} & & 7.7 & 11.4 & \multicolumn{1}{c|}{-} & 21.2 & 27.5 & \multicolumn{1}{c}{-} & & 6.8 & 10.8 & \multicolumn{1}{c}{-} \\

    TraCQ~\cite{desai2022single} \tnote{\textdaggerdbl} & \multicolumn{1}{c}{-} & \multicolumn{1}{c}{-} & \multicolumn{1}{c}{-} & & \multicolumn{1}{c}{-} & \multicolumn{1}{c}{-} & \multicolumn{1}{c|}{-} & \multicolumn{1}{c}{-} & \multicolumn{1}{c}{-} & \multicolumn{1}{c}{-} & & \multicolumn{1}{c}{-} & \multicolumn{1}{c}{-} & \multicolumn{1}{c|}{-} & 19.7 & 28.3 & \underline{35.7} & & 12.0 & 13.8 & 14.6 \\ 
    
    SGTR~\cite{li2022sgtr}  \tnote{\textsection} & \multicolumn{1}{c}{-} & \multicolumn{1}{c}{-} & \multicolumn{1}{c}{-} & & \multicolumn{1}{c}{-} & \multicolumn{1}{c}{-} & \multicolumn{1}{c|}{-} & \multicolumn{1}{c}{-} & \multicolumn{1}{c}{-} & \multicolumn{1}{c}{-} & & \multicolumn{1}{c}{-} & \multicolumn{1}{c}{-} & \multicolumn{1}{c|}{-} & \multicolumn{1}{c}{-} & 24.6 & 28.4 & & \multicolumn{1}{c}{-} & 12.0 & 15.2 \\ 

    SGTR~\cite{li2022sgtr,li2021bipartite} \tnote{\textsection}\enspace\tnote{*}  & \multicolumn{1}{c}{-} & \multicolumn{1}{c}{-} & \multicolumn{1}{c}{-} & & \multicolumn{1}{c}{-} & \multicolumn{1}{c}{-} & \multicolumn{1}{c|}{-} & \multicolumn{1}{c}{-} & \multicolumn{1}{c}{-} & \multicolumn{1}{c}{-} & & \multicolumn{1}{c}{-} & \multicolumn{1}{c}{-} & \multicolumn{1}{c|}{-} & \multicolumn{1}{c}{-} & 20.6 & 25.0 & & \multicolumn{1}{c}{-} & 15.8 & \underline{20.1} \\ 

    \midrule

    VCTree~\cite{tang2019learning} \tnote{\textbardbl}  & \underline{60.1} & \textbf{66.4} & \textbf{68.1} & & \multicolumn{1}{c}{-} & \multicolumn{1}{c}{-} & \multicolumn{1}{c|}{-} & \textbf{35.2} & \underline{38.1} & \underline{38.8} & & \multicolumn{1}{c}{-} & \multicolumn{1}{c}{-} & \multicolumn{1}{c|}{-} & \underline{22.0} & 27.9 & 31.3 & & \multicolumn{1}{c}{-} & \multicolumn{1}{c}{-} & \multicolumn{1}{c}{-} \\
    
    BGNN~\cite{li2021bipartite} \tnote{\textdagger}\enspace\tnote{*}  & \multicolumn{1}{c}{-} & 59.2 & 61.3 & & \multicolumn{1}{c}{-} & 30.4 & 32.9 & \multicolumn{1}{c}{-} & 37.4 & 38.5 & & \multicolumn{1}{c}{-} & 14.3 & 16.5 & \multicolumn{1}{c}{-} & \textbf{31.0} & \textbf{35.8} & & \multicolumn{1}{c}{-} & 10.7 & 12.6 \\

    PPDL~\cite{li2022ppdl} \tnote{\textdagger}\enspace\tnote{*}   &  \multicolumn{1}{c}{-}  & 41.6  & 43.6 & & \multicolumn{1}{c}{-}  & 33.3  & 36.2 & \multicolumn{1}{c}{-}  & 24.8  & 26.2 & & \multicolumn{1}{c}{-}  & 20.2  & 22.0 & \multicolumn{1}{c}{-}  & 13.6  & 16.5 & & \multicolumn{1}{c}{-} & 12.2 & 14.4 \\
    
    PCPL~\cite{yan2020pcpl} \tnote{\textbardbl}\enspace\;\tnote{*} & \multicolumn{1}{c}{-}  & 50.8  & 52.6 & & \multicolumn{1}{c}{-} & 35.2 & 37.8 & \multicolumn{1}{c}{-}  & 27.6  & 28.4 & & \multicolumn{1}{c}{-} & 18.6 & 19.6 & \multicolumn{1}{c}{-}  & 14.6 & 18.6 & & \multicolumn{1}{c}{-} & 9.5 & 11.7 \\
    
    RTPB~\cite{chen2022resistance} \tnote{\textdagger}\enspace\tnote{*} & \multicolumn{1}{c}{-} & 45.6 & 47.5 & & \underline{30.3} & 36.2 & 38.1 & \multicolumn{1}{c}{-} & 24.5 & 25.5 & & \underline{19.1} & 21.8 & 22.8 & \multicolumn{1}{c}{-} & 19.7 & 23.4 & & \underline{12.7} & \underline{16.5} & 19.0 \\
    
    DT2-ACBS~\cite{desai2021learning} \tnote{\textsection}\enspace\tnote{*} & \multicolumn{1}{c}{-} & 23.3 & 25.6 & & 27.4 & 35.9 & 39.7 & \multicolumn{1}{c}{-} & 16.2 & 17.6 & & 18.7 & \underline{24.8} & 27.5 & \multicolumn{1}{c}{-} & 15.0 & 16.3 & & \textbf{16.7} & \textbf{22.0} & \textbf{24.4} \\
    
    IETrans~\cite{zhang2022fine} \tnote{\textdagger}\enspace\tnote{*} & \multicolumn{1}{c}{-} & 48.0 & 49.9 & & \multicolumn{1}{c}{-} & 37.0 & 39.7 & \multicolumn{1}{c}{-} & 30.0 & 30.9 & & \multicolumn{1}{c}{-} & 19.9 & 21.8 & \multicolumn{1}{c}{-} & 23.6 & 27.8 & & \multicolumn{1}{c}{-} & 12.0 & 14.9 \\
    
    FGPL~\cite{lyu2022fine} \tnote{\textdagger}\enspace\tnote{*} & \multicolumn{1}{c}{-} & \multicolumn{1}{c}{-} & \multicolumn{1}{c}{-} & & \textbf{30.8} & \underline{37.5} & \underline{40.2} & \multicolumn{1}{c}{-} & \multicolumn{1}{c}{-} & \multicolumn{1}{c}{-} & & \textbf{21.9} & \textbf{26.2} & \underline{27.6} & \multicolumn{1}{c}{-} & \multicolumn{1}{c}{-} & \multicolumn{1}{c}{-} & & 11.9 & 16.2 & 19.1 \\

    \midrule
    RepSGG \tnote{\textdaggerdbl} & 55.1 & 62.8 & \underline{65.3} & & 17.1 & 22.2 & 24.4 & \underline{34.3} & \textbf{43.5} & \textbf{49.4} & & 10.1 & 13.8 & 16.9 & \textbf{22.5} & \underline{29.6} & 34.8 & & 6.7 & 9.3 & 11.4 \\
    RepSGG\textsubscript{PGLA} \tnote{\textdaggerdbl}\enspace\tnote{*}  & 24.3 & 27.8 & 28.8 & & 29.2 & \textbf{39.7} & \textbf{43.7} & 13.8 & 17.9 & 20.3 & & 16.2 & 22.3 & \textbf{27.7} & 8.7 & 12.1 & 14.6 & & 10.9 & 15.3 & 18.9 \\
    
    \bottomrule
    \end{tabular}
    \begin{tablenotes}
    Backbone network:
      \footnotesize
      \item [\textdagger] ResNeXt-101-FPN 
      \item [\textsection] ResNet-101 
      \item [\textbardbl] VGG-16
      \item [\textdaggerdbl] ResNet-50 
    \item[]{\hfill * debiasing technique is used}
    \end{tablenotes}
\end{threeparttable}
}
\end{table*}

\subsection{Datasets and Evaluation}

\textbf{Datasets}. We evaluate our methods on the Visual Genome (VG) \cite{krishna2017visual} dataset. We follow the protocols for the widely-used pre-processed subset VG150 \cite{xu2017scene} which contains the most frequent 150 entities ($C$ = 150) and 50 predicates ($P$ = 50). The dataset contains approximately 108k images, with 70\% for training and 30\% for testing. We also evaluate on the Open Images V6 dataset (OIV6)~\cite{kuznetsova2020open}, which contains 126k training images, 5k testing images, 301 entities and 30 predicates. For OIV6, as some entity and predicate classes are absent from the testing set, we exclusively train on the classes that are actually present. Consequently, we use 212 entities and 21 predicates that remain in the training set.

\textbf{Evaluation}. We evaluate our methods following three standard evaluation tasks:
\begin{enumerate*}[label=\arabic*)]
  \item predicate classification (PredCls): predict predicates given ground-truth entity classes and bounding boxes;
  \item scene graph classification (SGCls): predict predicates and entity classes given ground-truth entity bounding boxes;
  \item scene graph detection (SGDet): predict predicates, entity classes, and entity bounding boxes.
\end{enumerate*}
For VG150, we report results of recall@K (R@K)~\cite{lu2016visual}, mean recall@K (mR@K)~\cite{chen2019knowledge,tang2019learning}, zero-shot recall@K (zs-R@K)~\cite{lu2016visual} for all the three evaluation tasks. In addition, we further evaluate zero-shot mean recall@K (zs-mR@K) to assess methods' ability to generalize to unseen long-tailed testing distributions. For OIV6, following previous works~\cite{zhang2019graphical,li2021bipartite}, we report results of R@K, weighted mean AP of relationship detection (wmAP$_{rel}$), weighted mean AP of phrase detection (wmAP$_{phr}$), and the weighted score as score$_{wtd}$=$ 0.2 \times \text{R@50} + 0.4 \times \text{wmAP}_{rel} + 0.4 \times \text{wmAP}_{phr} $. Considering the down-weighting effect of these metrics on tailed predicates, we also provide mean recall@K results.

\subsection{Implementation Details} \label{sec:impl_details}

Multi-scale training is adopted following FCOS~\cite{tian2020fcos}. Images are resized such that their shorter edge is sampled from [480, 800] with a step of 32, and their longer edge does not exceed 1333 pixels. Random horizontal flip with a probability of 0.5 and random crop for are used for data augmentations. Specifically, a relative random ratio between 0.8 and 1 is selected to crop along each axis respectively. The repeat factor sampling \cite{gupta2019lvis} with the factor of 0.02 is applied to sample more images that contain tail predicates.
ResNet-50~\cite{he2016deep} is used as the backbone network and the same hyper-parameters are used following~\cite{tian2020fcos}. The entity detector is initialized with the weights pre-trained on COCO dataset~\cite{lin2014microsoft}, and trained for 90k iterations. The relation detection modules are trained for additional 90k iterations while freezing the backbone and entity detection heads. Finally, the entire model is jointly trained for 10k iterations, and this procedure is referred to as fine-tuning throughout the rest of the paper.
Training is performed on 4 Nvidia A100 GPUs with a batch size of 32.
The AdamW~\cite{loshchilov2018decoupled} optimizer is used with a initial learning rate of $10^{-5}$ which is decayed at the 80k-th iteration by a factor of 0.1, and the weight decay of $10^{-4}$. The learning rates of the backbone and rep-point samplers are multiplied by a factor of 0.1. A single model is trained for all tasks, rather than separate models for each task. For the encoder, $L_e$ is set to 1 with the same hyper-parameter setting as used in~\cite{zhu2021deformable}. The decoder is configured with $L_d = 1$, $K=4$, $h_G = h_R = 8$, $h_A = 128$, $d_G = d_R = 32$, and $d_A = 64$ as the default settings. For the rep-point samplers, $m$ is uniformly sampled from [1, 100].

\subsection{Quantitative Results}

\begin{figure*}[t]
\centering
\pgfplotstableread[col sep=comma]{data/head.csv}\head
\pgfplotstableread[col sep=comma]{data/body.csv}\body
\pgfplotstableread[col sep=comma]{data/tail.csv}\tail
\definecolor{methodA}{HTML}{E3C75F}
\definecolor{methodB}{HTML}{FF4858}
\definecolor{methodC}{HTML}{80A6E2}
\newcolumntype{C}[1]{>{\centering\arraybackslash }b{#1}}
\pgfplotstableread[col sep=space, trim cells]{
Empty {RepSGG} {RepSGG2} {FGPL}
{head} 31.2857143	15.2857143	24.6685714
{body} 11.2380952	21.7619048	21.2385714
{tail} 5.2727273	17.2727273	13.2409091
}\MyTable

\begin{tikzpicture}
\pgfplotsset{every axis title/.append style={at={(-0.006,0.75)}, anchor=south west}}
\begin{groupplot}[group style={group size=1 by 3},
    ybar=0pt,
    ymin=0,
    ymax=0.7,
    xtick=data,
    xlabel={},
    xticklabel style={rotate=45, anchor=north east},
    ytick={0,0.2,0.4,0.6},
    yticklabels={0,0.2,0.4,0.6},
    xticklabel style={yshift=5pt, xshift=5pt},
    width=\linewidth,
    height=0.15\linewidth,
    enlarge x limits={abs=0.5},
    xticklabel style={font=\scriptsize},
    yticklabel style={font=\scriptsize},
    xtick pos=bottom,
    ytick pos=left,
    yticklabel = {\pgfmathparse{\tick*100}\pgfmathprintnumber{\pgfmathresult}}
    ]
    
    \nextgroupplot[title=head,bar width=5pt,
    xticklabels from table={\head}{method},
    legend style={column sep=5pt, fill=white, fill opacity=1.0, draw opacity=1, text opacity=1,nodes={scale=1.0, transform shape}, at={(0.7,0.95)}},
    legend cell align={left}, legend columns=3]
    \addplot[fill=methodA, draw=methodA]
    table[x expr=\coordindex, y=baseline, col sep=comma] {\head};
    \addplot[fill=methodB, draw=methodB]
    table[x expr=\coordindex, y=repsgg, col sep=comma] {\head};
    \addplot[fill=methodC, draw=methodC]
    table[x expr=\coordindex, y=fgpl, col sep=comma] {\head}; 
    \legend{RepSGG, RepSGG\textsubscript{PGLA}, FGPL}
    
    \nextgroupplot[ylabel={Recall@100},title=body, bar width=5pt,xticklabels from table={\body}{method}]
    \addplot[fill=methodA, draw=methodA]
    table[x expr=\coordindex, y=baseline, col sep=comma] {\body};
    \addplot[fill=methodB, draw=methodB]
    table[x expr=\coordindex, y=repsgg, col sep=comma] {\body};
    \addplot[fill=methodC, draw=methodC]
    table[x expr=\coordindex, y=fgpl, col sep=comma] {\body}; 

    \nextgroupplot[title=tail, bar width=5pt,xticklabels from table={\tail}{method}]
    \addplot[fill=methodA, draw=methodA]
    table[x expr=\coordindex, y=baseline, col sep=comma] {\tail};
    \addplot[fill=methodB, draw=methodB]
    table[x expr=\coordindex, y=repsgg, col sep=comma] {\tail};
    \addplot[fill=methodC, draw=methodC]
    table[x expr=\coordindex, y=fgpl, col sep=comma] {\tail};

\end{groupplot}

\node[above left,
      text width=6.5cm,
      minimum height=2.8cm,
      fill=white,
      align=center, scale=0.7] (groupedtable) at (16.7, -0.535) {%
    
    \pgfplotstabletypeset[
        fixed,
        precision=1,
        fixed zerofill,
        col sep=space,
        every head row/.style={
                before row={\toprule},
                after row={\midrule}
            },
        every last row/.style={after row=\bottomrule},
        every first column/.style={column type={l|},string type},
        columns/Empty/.style={string type,column name={}},
        columns/RepSGG2/.append style={column name={RepSGG\textsubscript{PGLA}}},
        every row 0 column 1/.style={postproc cell content/.style={
          @cell content/.add={$\bf}{$}
        }},
        every row 1 column 2/.style={postproc cell content/.style={
          @cell content/.add={$\bf}{$}
        }},
        every row 2 column 2/.style={postproc cell content/.style={
          @cell content/.add={$\bf}{$}
        }},
    ]\MyTable};

\node[inner sep=0pt, above=-0.8\abovecaptionskip, scale=0.9] at (groupedtable.north) {Group R@100.};

\end{tikzpicture}
\caption{Per-predicate SGDet R@100 comparison between RepSGG, RepSGG\textsubscript{PGLA}, and FGPL on VG150 dataset. RepSGG\textsubscript{PGLA} performs better on body and tail groups. The overall standard deviation of R@100 is 14.6 (RepSGG), 12.3 (RepSGG\textsubscript{PGLA}), and 13.6 (FGPL) respectively, which also implies that RepSGG\textsubscript{PGLA} achieves a more balanced performance.}
\label{fig:perclass}
\end{figure*}
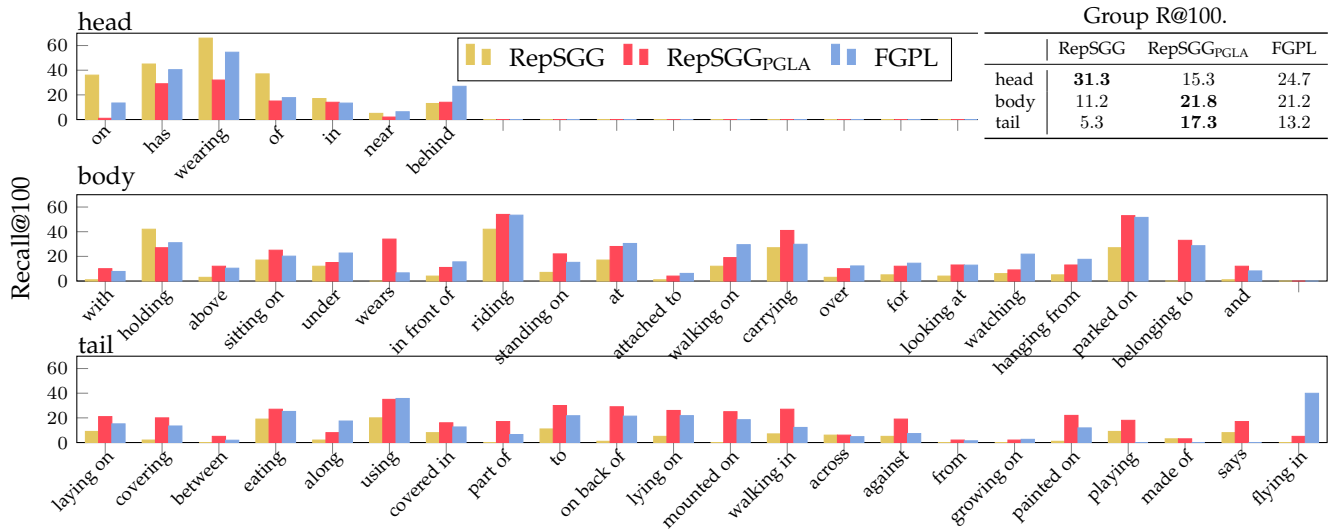

\subsubsection{Visual Genome}

To compare with methods using different entity representations and those using debiasing techniques respectively, we train one model without PGLA and another with PGLA. R@K is the main metric when comparing methods without debiasing, while mR@K is the main one for debiasing methods. 
We compare our methods with the state-of-the-art (SOTA) scene graph generation models as shown in Table~\ref{tab:sota}.
We divide the methods for comparison into 3 groups in Table~\ref{tab:sota} from top to bottom: point-based, query-based, and box-based, regardless of whether long-tailed techniques are used. 
Notably, point-based and query-based methods seek fast inference speed, and often do not involve debiasing techniques. While box-based methods focus on designing debiasing methods with off-the-shelf entity detectors.

Solely comparing methods that have different entity or predicate representations, RepSGG outperforms point-based methods FCSGG~\cite{liu2021fully} and CoRF~\cite{Adaimi_2023_WACV} on all metrics by a large margin. 
Comparing with query-based methods, RepSGG outperforms RelTR~\cite{cong2023reltr}, TraCQ~\cite{desai2022single}, and SGTR~\cite{li2022sgtr} on most recall metrics across all 3 tasks. 
Under the condition of no debiasing, the performance improvements on R@K indicates that the proposed entity and predicate representations are superior to point-based and query-based methods.
Compared with box-based VCTree~\cite{tang2019learning}, RepSGG achieves higher R@K on SGCls and SGDet tasks with slightly lower recall on PredCls.
In terms of debiasing methods, RepSGG\textsubscript{PGLA} outperforms the SOTA methods on PredCls mR@50, PredCls mR@100, and SGCls mR@100, while achieving comparable results on other metrics. RepSGG achieves 43.7 mR@100 on PredCls, which is 3.5 higher than the box-based FGPL~\cite{lyu2022fine}. FGPL is a complicated long-tail learning technique which requires a biased model with several fine-tuned hyper-parameters, while PGLA is much simpler yet effective for mitigating the long-tailed problem. 

We further compare the per-predicate and group R@100 with FGPL for the SGDet task as shown in Fig.~\ref{fig:perclass}. 
Predicates are sorted in descending order based on their frequency in the training set, and divided into 3 groups following~\cite{li2021bipartite}.
RepSGG\textsubscript{PGLA} achieves higher recall on body and tail classes, resulting in a more balanced performance over FGPL. In VG150 testing set, there are only 29, 37, and 12 triplets involving \texttt{playing}, \texttt{made of}, and \texttt{says}, respectively. While FGPL failed retrieving these triplets, our method achieves significantly better results, even thought these triplets are extremely rare both during training and testing. Notably, without debiasing techniques, RepSGG achieves excellent performance on some tail predicates as well, such as \texttt{playing}, \texttt{made of}, and \texttt{says}.
It demonstrates that the proposed entity and relationship representations inherently capture more informative semantics. 

\begin{table}[t]
\caption{PredCls results of zero-shot mean recall (zs-mR@K) and zero-shot recall (zs-R@K) on VG150 compared to state-of-the-art methods. The best results are bold, and the second-best results are underlined.}
\label{tab:zeroshot}
\centering
\sisetup{table-format=2.1, table-number-alignment=right, table-space-text-post=\hspace{0.0cm}, mode=text, detect-weight}
\renewcommand{\arraystretch}{1.1}
\begin{tabular}{l| S S S| S S S } 
\toprule
  & \multicolumn{6}{c} {PredCls Zero Shot Relationship Retrieval}  \\ 
& \multicolumn{3}{c|}{mR@20/50/100} & \multicolumn{3}{c}{R@20/50/100} \\
\midrule
BGNN~\cite{li2021bipartite} & 1.9 & 3.2  & 4.9 & 2.0 & 3.5 & 4.6   \\
BA-SGG~\cite{guo2021general} & 3.1  & 5.3 & 6.7 & 3.0 & 6.0 & 8.0   \\
Motifs-TDE~\cite{tang2020unbiased} & 5.3 & 9.3 & 11.4 & 8.3  & \Uline{14.3}  & \ubold 18.0 \\
FGPL~\cite{lyu2022fine} & \ubold 11.0  & 14.3  & 15.9 & \ubold 9.4 & 13.0 & 14.6   \\
IETrans~\cite{zhang2022fine} & \ubold 11.0 & \Uline{14.5} & \Uline{17.0} & 6.5 & 10.0 & 12.0 \\
\midrule
RepSGG & 4.3 & 7.1 & 8.6 & \Uline{8.7} & \ubold 14.6 & \Uline{17.9} \\
RepSGG\textsubscript{PGLA} & \Uline{9.9} & \ubold 17.2 & \ubold 20.1 & 6.1 & 9.2 & 11.0 \\
\bottomrule
\end{tabular}
\end{table}

We further conduct analysis on zero-shot performance. In this setting, the objective is to retrieve triplets that are not encountered during training, but are present during testing. 
In Table~\ref{tab:zeroshot}, we report the zero-shot recall and zero-shot mean recall results on the PredCls task and compare with the SOTA methods. We collect the results by implementing the zs-mR@K evaluation using the corresponding open-sourced codes.
RepSGG\textsubscript{PGLA} outperform the SOTA methods on mR@50 and mR@100 by a large margin. Among methods for comparison, IETrans~\cite{zhang2022fine} achieves good results by re-labeling predictions and labeling unannotated samples from biased models for training. Without extra data for training, RepSGG\textsubscript{PGLA} outperforms IETrans by 3.1 on zs-mR@100. RepSGG also achieves high zero-shot recall comparable to the SOTA methods.
It demonstrates that RepSGG generalizes significantly better to compositions of entities and relationships in unseen contexts.

\subsubsection{Open Images}

\begin{table}[t]
\caption{Comparisons with the state-of-the-art methods on OI V6. R@50 in the table is micro-Recall@50~\cite{gkanatsios2019attention}. The best results are bold, and the second-best results are underlined.}
\label{tab:oiv6}
\resizebox{\linewidth}{!}{%
\centering
\begin{tabular}{l|ccccc}
\toprule  
& mR@50 & R@50 & wmAP$_{rel}$ & wmAP$_{phr}$ & score$_{wtd}$ \\
\midrule
Motifs~\cite{zellers2018neural} & 32.68 & 71.63 & 29.91 & 31.59 & 38.93 \\
RelDN~\cite{zhang2019graphical} & 33.98 & 73.08 & 32.16 & 33.39 & 40.84 \\
VCTree~\cite{tang2019learning} & 33.91 & 74.08 & 34.16 & 33.11 & 40.21 \\
G-RCNN~\cite{yang2018graph} & 34.04 & 74.51 & 33.15 & 34.21 & 41.84 \\
GPS-Net~\cite{lin2020gps} & 35.26 & 74.81 & 32.85 & 33.98 & 41.69 \\
BGNN~\cite{li2021bipartite} & 40.45 & 74.98 & 33.51 & 34.15 & 42.06 \\
SGTR~\cite{li2022sgtr} & 42.61 & 59.91 & \underline{38.73} & \underline{36.98} & 42.28 \\
SS R-CNN~\cite{teng2022structured} & 50.73 & \textbf{75.70} & \textbf{41.14} & \textbf{43.24} & \textbf{48.89} \\
CSL~\cite{liu2023constrained} & 41.72 & \underline{75.44} & 34.30 & 35.38 & \underline{42.86} \\
\midrule
RepSGG& \underline{56.65} & 73.27 & 19.10 & 19.50 & 30.10 \\
RepSGG\textsubscript{PGLA/R} & \textbf{59.82} & 66.52 & 24.34 & 24.58 & 32.87 \\
RepSGG\textsubscript{PGLA/P} & 52.44 & 65.10 & 27.28 & 27.50 & 34.93 \\

\bottomrule
\end{tabular}
}
\end{table}
Since the precision is one of the evaluation metrics for OIV6, we conduct experiments using precision as the PGLA metric besides recall.
The model trained with precision-guided logit adjustment is denoted as RepSGG\textsubscript{PGLA/P}, and the model trained with recall-guided logit adjustment is renamed to RepSGG\textsubscript{PGLA/R}.
The experimental results on OIV6~\cite{kuznetsova2020open} are shown in Table~\ref{tab:oiv6}.
We observe that all RepSGG models outperforms other methods on mR@50 by a large margin. RepSGG\textsubscript{PGLA/R} achieves a mR@50 of 59.82, which is 9.09 higher than SS R-CNN~\cite{teng2022structured}.
RepSGG\textsubscript{PGLA/P} achieves better results on wmAP$_{rel}$ and wmAP$_{phr}$ in comparison to RepSGG and RepSGG\textsubscript{PGLA/R}, highlighting the effectiveness of precision-guided PGLA for precision-oriented tasks.
Our methods achieves lower precision-oriented metrics like wmAP$_{rel}$, wmAP$_{phr}$, and score$_{wtd}$.
This is because OIV6 has very sparse annotations and our model has great generalization power as shown in Table~\ref{tab:zeroshot}. OIV6 has 2.76 relationship annotations per image on average in the training set, while VG150 has 5.97. As a result, most detections will be considered as false positives, leading to lower precision.
We then collect the results on other SGG tasks to further analyze the performance as shown in Table~\ref{tab:oi_recall}.
The results on R@50, mR@50, and wmAP$_{rel}$ show significant improvements on PredCls and SGCls tasks.
In the PredCls setting, all RepSGG models achieve R@50 and wmAP$_{rel}$ over 90.
This highlights that the bottleneck of the model is primarily in entity detection, rather than predicate prediction.

\begin{table}[t]
\centering
\caption{Comparisons of RepSGG models without PGLA, with recall-guided LA, and with precision-guided LA on PredCls and SGCls tasks.}
\label{tab:oi_recall}
\resizebox{\linewidth}{!}{%

\sisetup{table-format=2.1, table-number-alignment=center, table-space-text-post=\hspace{0pt}}
\setlength{\tabcolsep}{4pt}
\renewcommand{\arraystretch}{1.1}
    \begin{tabular}{l | 
    SSS | 
    SSS }
    \toprule 
      & \multicolumn{3}{c|} {PredCls} & \multicolumn{3}{c} { SGCls }\\ 
    & \multicolumn{1}{c}{mR@50} & \multicolumn{1}{c}{R@50} & \multicolumn{1}{c|}{wmAP$_{rel}$} & \multicolumn{1}{c}{mR@50} & \multicolumn{1}{c}{R@50} & \multicolumn{1}{c}{wmAP$_{rel}$} \\
    \midrule
    RepSGG  & 70.12 & 97.93 & 93.17 & 64.70 & 87.06  & 36.80  \\
    RepSGG\textsubscript{PGLA/R}  & 79.03 & 97.05 & 90.50 &  69.24 & 80.51 & 41.47  \\
    RepSGG\textsubscript{PGLA/P}  & 76.99 & 95.37 & 92.29 & 57.79 & 74.84 &  43.01 \\
    
    \bottomrule
    \end{tabular}
}

\end{table}


    


\subsection{Qualitative Analysis}

\begin{figure}[t]
\centering
\includegraphics[width=\linewidth]{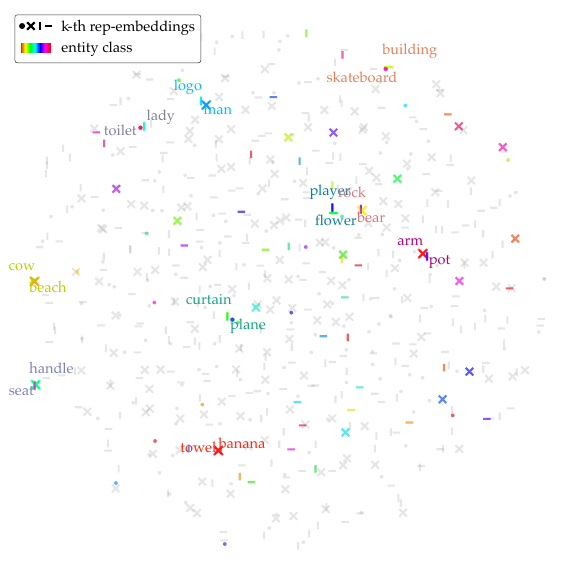}
\caption{The t-SNE visualization results on trained subject rep-embeddings. The pairwise cosine similarity is used, and the top-10 pairs are labeled. Rep-embeddings of an entity class have the same color. Only the involved entity classes in the top-10 pairs are colored, while other classes are in gray.}
\label{fig:init_sbj_embeds}
\end{figure}

\begin{figure}[t]
\centering
\includegraphics[width=\linewidth]{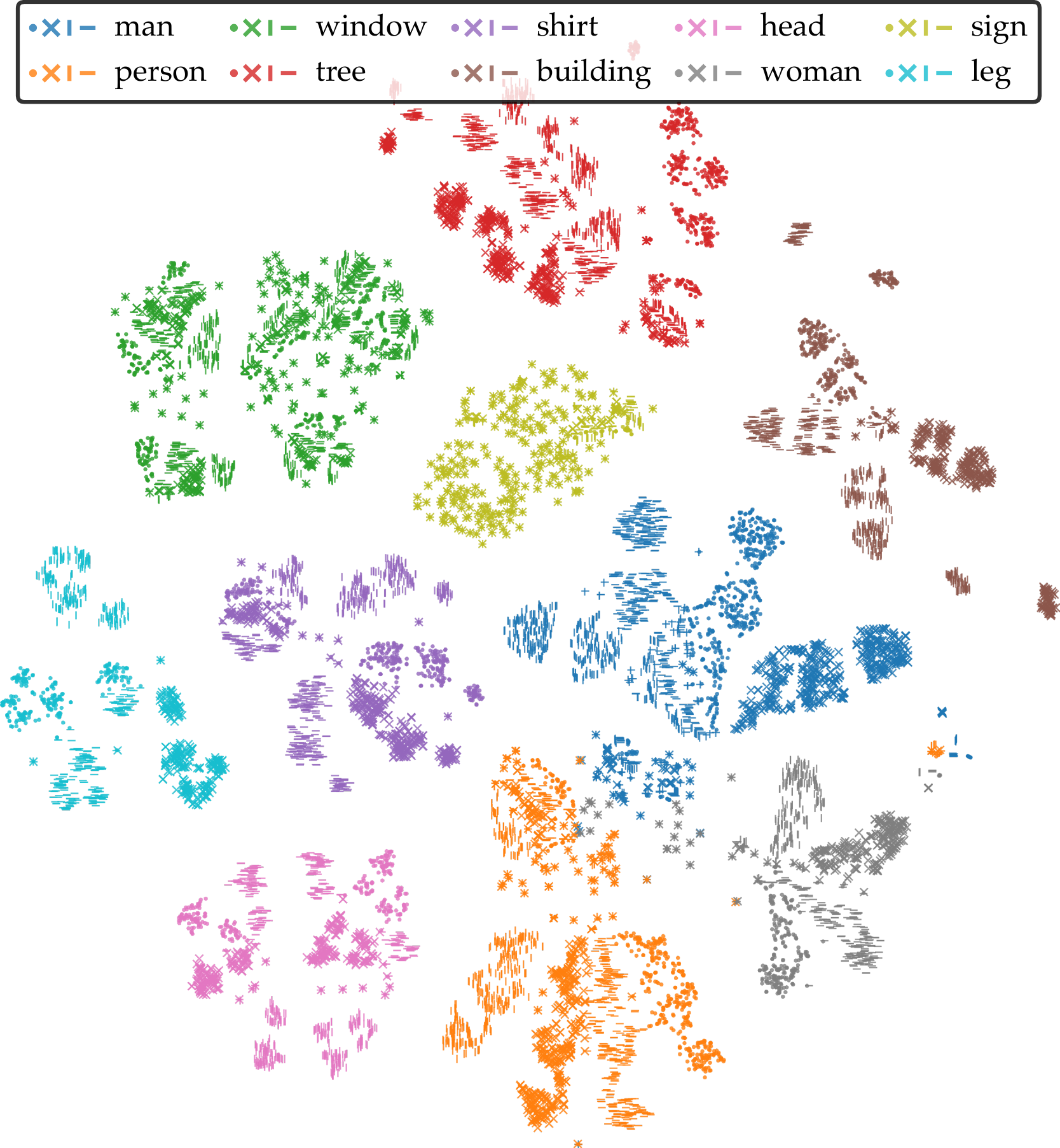}
\caption{The t-SNE visualization results on the output subject queries of the decoder ($\rmQ^L$) for 10 frequent entity classes.}
\label{fig:final_sbj_embeds}
\end{figure}

\begin{figure*}[t]
\centering
\includegraphics[width=\linewidth]{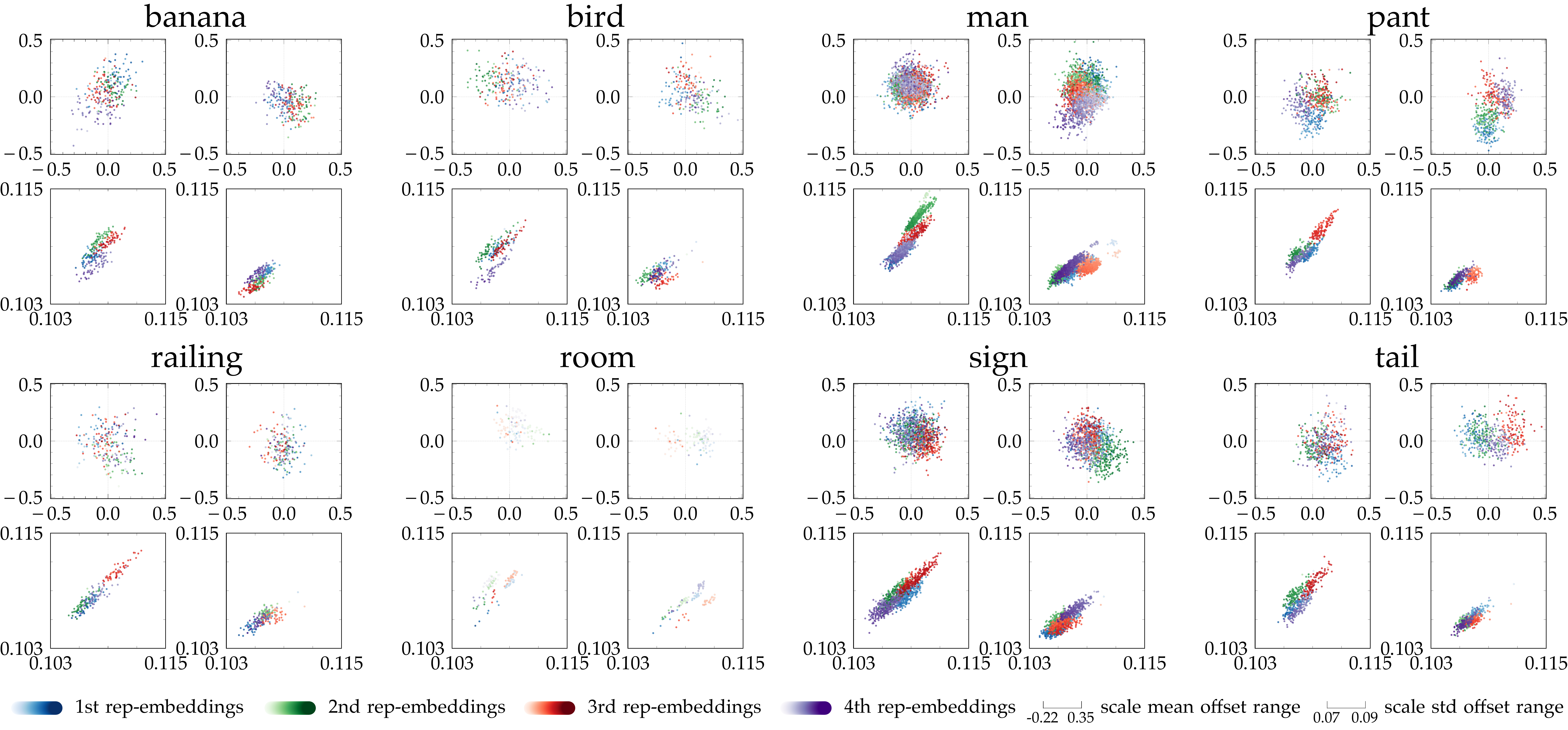}
\caption{Visualizations on predicted subject and object rep-point mean and std offsets. For each entity class, the offsets of subject means, object means, subject stds, and object stds are shown on the top-left, top-right, bottom-left, and bottom-right of the sub-figure. The coordinates represent spatial offsets \wrt the ground-truth centers, while the color saturation denotes the scale offsets \wrt the entity feature scale. The more saturated the color is, the larger the scale offset is, and vice versa.}
\label{fig:offsets}
\end{figure*}

We want to qualitatively examine what the model learns from the data, especially on the rep-embeddings and rep-points. We visualize the weights of subject rep-embeddings $\rmE_s$ of the trained RepSGG\textsubscript{PGLA} (with $K=4$, $L_e=1$, and $L_d=1$) via t-SNE~\cite{van2008visualizing} as shown in Fig.~\ref{fig:init_sbj_embeds}.
The top 10 pairs of similar rep-embeddings are highlighted, while the remaining rep-embeddings belonging to the involved entities are colored.
We use the pairwise cosine similarity as the distance metric for t-SNE where distances represent the similarities between rep-embeddings. 
It reveals that the rep-embeddings between different entity classes are well separated. 
Rep-embeddings of the same entity class are distinctly separated as well.
Interestingly, most pairs do not exhibit explicit semantic part affinities.
We hypothesize that the initial rep-embeddings serve as ``anchors'' which are evenly distributed in the embedding space, and capture more semantic information as they progress through the decoder.

We further validate the hypothesis by visualizing the output queries $\rmQ^L$.
To eliminate entity mis-classification, we collect the output queries on the PredCls task from the first 1000 testing images.
As shown in Fig.~\ref{fig:final_sbj_embeds}, the inter-class rep-embeddings are distinguishable.
Entities with similar semantics, such as \texttt{woman}, \texttt{man}, and \texttt{person}, form more compact clusters.
The compactness and distinctness of rep-embeddings differ within an entity class.
For \texttt{building}, 4 types of rep-embeddings are well-separated with high variance where each type is responsible for a different semantic concept.
For \texttt{sign}, the distribution of rep-embeddings is more compact, indicating that the semantics and relationships associated with \texttt{sign} are mostly homogeneous.

We also collect the subject and object rep-point offset parameters $\bm{\mu}^1_s$, $\bm{\sigma}^1_s$, $\bm{\mu}^1_o$, and $\bm{\sigma}^1_o$, which are shown in Fig.~\ref{fig:offsets}.
The variation in parameters is evident, primarily between different entity classes, and between subject and object offsets.
The distribution of parameters implies the locations of relevant semantic features to a certain extent.
It is likely that the spatial means follow a bivariate Gaussian distribution especially for \texttt{man} and \texttt{sign}, which suggests that rep-points are sampled around the bounding box centers.
Entities with larger location variances within bounding boxes, such as \texttt{banana}, \texttt{bird}, and \texttt{railing}, exhibit more dispersed distributions of spatial mean offsets.
The distributions of subject and object mean offsets for \texttt{man} are noticeably distinct.
The subject means are more tightly clustered around centers, indicating that when \texttt{man} is a subject, the features representing the entirety of a \texttt{man} are more significant.
Conversely, when \texttt{man} is an object, the features representing semantic parts become more important as object means are much more distributed.
The scale mean offsets also reveal different patterns among entities.
For entities with large bounding boxes like \texttt{room}, the scale means are primarily negative, indicating that the sampled rep-points are from lower scales of features.
For entities in medium and small sizes, the sampled rep-points are mostly from the same or higher scales of features.
In terms of the standard deviations (stds) of offsets, we observe the spatial collinearity, which is a natural occurrence in training data with diverse entity sizes.
It is notable that \texttt{railing} has larger spatial and scale stds, reflecting the uncertainty associated with the varying lengths of railings.
The scale stds normally remain within a narrower range (0.07 to 0.09) compared to the spatial stds (0.103 to 0.115).

\subsection{Ablation Studies}

To further investigate the proposed methods, we perform ablation studies on VG150 dataset. Unless specified, we use the same hyper-parameters as discussed in Section~\ref{sec:impl_details}, and PGLA is applied but fine-tuning is not applied.

\subsubsection{Analysis of RepSGG}

\begin{table}[t]
\caption{Ablation studies of number of rep-embeddings $K$, number of encoder layers $L_e$, and number of decoder layers $L_d$ in RepSGG\textsubscript{PGLA}. Results on mR@100 and zs-mR@100 are collected for three SGG tasks.}
\label{tab:ablation_layers}
\resizebox{\linewidth}{!}{%
\centering
\renewcommand{\arraystretch}{1.1}
\sisetup{table-format=2.1, table-number-alignment=center, table-space-text-post=\hspace{0.0cm}}
\begin{tabular}{r c c | S S | S S | S S} 
\toprule 
  \multirow{2}{*}{$K$} & \multirow{2}{*}{$L_e$} & \multirow{2}{*}{$L_d$} & \multicolumn{2}{c|}{PredCls @100} & \multicolumn{2}{c|}{SGCls @100} & \multicolumn{2}{c}{SGDet @100} \\
& & & \multicolumn{1}{c}{mR} & \multicolumn{1}{c|}{zs-mR} & \multicolumn{1}{c}{mR} & \multicolumn{1}{c|}{zs-mR} & \multicolumn{1}{c}{mR} & \multicolumn{1}{c}{zs-mR}\\
\midrule

1 & 1 & 1 & 41.0 & 18.6 & 21.5 & 4.9 & 15.4 & 2.6 \\
4 & 1 & 1 & 41.2 & 19.9 & 21.6 & 4.8 & 15.4 & 2.8 \\
7 & 1 & 1 & 41.5 & 18.7 & 21.6 & 4.8 & 15.3 & 3.3 \\
10 & 1 & 1 & 41.2 & 20.1 & 21.4 & 5.6 & 15.3 & 3.6 \\
4 & 0 & 0 & 39.3 & 15.8 & 19.9 & 4.7 & 13.6 & 1.8 \\
4 & 1 & 0 & 40.1 & 17.3 & 20.6 & 4.3 & 13.1 & 2.0 \\
4 & 0 & 1 & 40.9 & 17.6 & 20.8 & 4.9 & 15.4 & 2.6 \\
4 & 2 & 2 & 41.2 & 18.4 & 21.3 & 5.5 & 14.7 & 2.9 \\
10 & 3 & 3 & 41.4 & 19.6 & 21.8 & 5.4 & 14.2 & 2.8 \\

\bottomrule
\end{tabular}
}
\end{table}

\begin{table}[t]
\caption{Ablation studies of GCA and RCA.}
\label{tab:decoder}
\centering
\setlength{\tabcolsep}{3pt}
\renewcommand{\arraystretch}{1.1}
\sisetup{table-format=2.1, table-number-alignment=center, table-space-text-post=\hspace{0.0cm}}
\begin{tabular}{c c | S S | S S | S S} 
\toprule 
  \multicolumn{2}{c|}{} & \multicolumn{2}{c|}{PredCls @100} & \multicolumn{2}{c|}{SGCls @100} & \multicolumn{2}{c}{SGDet @100} \\
GCA & RCA & \multicolumn{1}{c}{mR} & \multicolumn{1}{c|}{zs-mR} & \multicolumn{1}{c}{mR} & \multicolumn{1}{c|}{zs-mR} & \multicolumn{1}{c}{mR} & \multicolumn{1}{c}{zs-mR}\\
\midrule
   $\checkmark$ & & 40.6 & 17.2 & 20.2 & 5.3 & 15.5 & 4.7 \\
     & $\checkmark$ & 41.4 & 19.8 & 21.2 & 6.5 & 15.5 & 3.7 \\
   $\checkmark$ & $\checkmark$ & 41.4 & 20.4 & 21.7 & 6.7 & 15.9 & 3.3 \\

\bottomrule
\end{tabular}
\end{table}
We investigate the effects of different numbers of rep-embeddings, encoder layers, and decoder layers. The results on mR@100 and zs-mR@100 are listed in Table~\ref{tab:ablation_layers}.
First, adding the encoder or decoder improves the overall performance.
It is also important to note that similar behaviors can be observed when increasing the values of $K$, $L_e$, or $L_d$, as the performance improves and then plateaus at certain values of $K$, $L_e$, or $L_d$, respectively.
The combination of $K=4$, $L_e=1$, and $L_d=1$ achieves a balanced trade-off between performance and inference speed.
Based on our experiments, four rep-embeddings per entity class ($K=4$) are sufficient to capture possible relationships, and using $K=10$ potentially causes overfitting on few re-embeddings.
Likewise, increasing the number of encoder or decoder layers does not lead to significant performance improvement.
When both encoder and decoder are removed, ReSGG\textsubscript{PGLA} still achieves a high PredCls mR@100 of 39.3, which outperforms many state-of-the-art box-based methods listed in Table~\ref{tab:sota}.
This suggests the advantage of our proposed relationship representation over traditional softmax classification.
Treating relationships as attention weights naturally incorporates more semantic information, effectively capturing the distinction between subject and object entities.

We further conduct the analysis of GCA and RCA modules in the decoder. We use the model with $K=4$, $L_e=1$, and $L_d=1$. We train 3 separate models, one with GCA only, one with RCA only, and one with both modules. When GCA is not equipped, we use the visual features sampled at entity centers as the inputs to RCA, without using rep-point samplers. The ablation results are shown in Table~\ref{tab:decoder}.
No significant performance drop is observed when removing GCA or RCA. RCA is more important than GCA based on the results, as the RCA-only model achieves better performance on most metrics. We conjecture that RCA exchanges semantic information among rep-embeddings, while GCA only exchanges visual features locally.
When both GCA and RCA are applied, we obtain the best results on most metrics except on SGDet zs-mR@100.

\subsubsection{Analysis of PGLA}

\begin{table}[t]
\caption{Ablation studies of loss and training configurations.}
\label{tab:losses}
\resizebox{\linewidth}{!}{%
\centering
\setlength{\tabcolsep}{3pt}
\renewcommand{\arraystretch}{1.1}
\sisetup{table-format=2.1, table-number-alignment=center, table-space-text-post=\hspace{0.0cm}}
\begin{tabular}{c c c c | S S | S S | S S} 
\toprule
  \multirow{2}{*}{LA} & \multirow{2}{*}{PGLA} & \multirow{2}{*}{$\mathcal{L}_{\eta}$} & \multirow{2}{*}{finetune} & \multicolumn{2}{c|}{PredCls @100} & \multicolumn{2}{c|}{SGCls @100} & \multicolumn{2}{c}{SGDet @100} \\
& & & & \multicolumn{1}{c}{mR} & \multicolumn{1}{c|}{zs-mR} & \multicolumn{1}{c}{mR} & \multicolumn{1}{c|}{zs-mR} & \multicolumn{1}{c}{mR} & \multicolumn{1}{c}{zs-mR}\\
\midrule
   & & & & 27.1 & 9.7 & 15.9 & 2.1 & 11.4 & 1.1 \\
   $\checkmark$ & & & & 39.4 & 16.2 & 19.4 & 4.5 & 13.7 & 2.4 \\
   & $\checkmark$ & & & 40.3 & 18.4 & 21.4 & 5.2 & 15.0 & 3.0 \\
   & $\checkmark$ & $\checkmark$ & & 41.2 & 20.1 & 21.4 & 5.6 & 15.3 & 3.6 \\
   & $\checkmark$ & $\checkmark$ & $\checkmark$ & 43.7 & 20.1 & 27.7 & 7.7 & 18.9 & 3.5 \\
\bottomrule
\end{tabular}
}
\end{table}

To investigate the effects different loss-related configurations proposed in the paper, we conduct the ablation experiments on Logit Adjustment~\cite{menon2021longtail}, the proposed PGLA, the margin ranking loss $\mathcal{L}_{\eta}$ proposed in Section~\ref{sec:losses}, and fine-tuning.
As shown in Table~\ref{tab:losses}, RepSGG\textsubscript{PGLA} achieves higher mean recall on all metrics compared with RepSGG trained with LA, which confirms the effectiveness of PGLA.
RepSGG\textsubscript{PGLA} also achieves a more significant improvement on zs-mR@100, providing evidence that PGLA is more resilient to under-fitting or over-fitting.
Using $\mathcal{L}_{\eta}$ with PGLA brings more improvements, as it effectively suppresses unlikely relationships and avoids excessively penalizing potentially unannotated ones.
Finally, with additional fine-tuning the entire model, we manage to further increase the performance on mean recall.

\begin{table}[t]
\caption{Effects of $\rmW$, $\rmB$, and $\rmD$ in PGLA.}
\label{tab:ablation_pgla}
\centering
\setlength{\tabcolsep}{3pt}
\renewcommand{\arraystretch}{1.1}
\sisetup{table-format=2.1, table-number-alignment=center, table-space-text-post=\hspace{0.0cm}}
\begin{tabular}{ c c c | S S | S S | S S} 
\toprule
  \multicolumn{3}{c|}{} & \multicolumn{2}{c|}{PredCls @100} & \multicolumn{2}{c|}{SGCls @100} & \multicolumn{2}{c}{SGDet @100} \\
$\rmW$ & $\rmB$ & $\rmD$ & \multicolumn{1}{c}{mR} & \multicolumn{1}{c|}{zs-mR} & \multicolumn{1}{c}{mR} & \multicolumn{1}{c|}{zs-mR} & \multicolumn{1}{c}{mR} & \multicolumn{1}{c}{zs-mR}\\
\midrule
   & & & 27.1 & 9.7 & 15.9 & 2.1 & 11.4 & 1.1 \\
   $\checkmark$ & & & 28.5 & 9.4 & 17.4 & 2.4 & 12.8 & 1.5 \\
   & $\checkmark$ & & 42.0 & 17.7 & 21.3 & 5.5 & 16.2 & 3.8 \\
   & & $\checkmark$ & 30.5 & 10.0 & 18.8 & 2.7 & 13.6 & 1.8 \\
   & $\checkmark$ & $\checkmark$ & 40.5 & 16.8 & 20.5 & 5.9 & 15.3 & 3.7 \\
   $\checkmark$ & & $\checkmark$ & 30.5 & 10.2 & 17.8 & 2.9 & 13.5 & 1.8 \\
   $\checkmark$ & $\checkmark$ & & 41.6 & 18.7 & 21.4 & 6.0 & 16.0 & 3.9 \\
   $\checkmark$ & $\checkmark$ & $\checkmark$ & 41.2 & 20.1 & 21.4 & 5.6 & 15.3 & 3.6 \\
\bottomrule
\end{tabular}
\end{table}

We study the individual and combinatorial effects of $\rmW$, $\rmB$, and $\rmD$ in (\ref{eq:la2}).
We simply set $\rmW = \bm{1}$ when $\rmW$ is not applied, and $\rmB$ or $\rmD$ to zero when they are not applied.
As shown in Table~\ref{tab:ablation_pgla}, introducing either component improves the performance on mean recall.
The bias term $\rmB$ has the most significant effect over the results with the largest improvement compared with individual application of $\rmW$ or $\rmD$. 
This can be attributed to the fact that class margins are primarily determined by the bias.
Similarly, $\rmD$ serves as an additional adjustment for class margins, which has more effect over $\rmW$ as a result.
Applying $\rmB$ alone already yields the highest mR, but considering zs-mR is also crucial for real applications of scene graphs.
Combing $\rmW$ and $\rmB$ enhances the zs-mR performance, and combing all 3 components further improves the zs-mR performance with a slightly decrease on mR.

\begin{figure}[t]
\centering
\pgfdeclareplotmark{mystar}{
    \node[star,star point ratio=2.25,minimum size=8pt,
          inner sep=0pt,draw=black,solid,fill=red] {};
}
\definecolor{mycolor}{RGB}{255,34,65}

\begin{tikzpicture}
\pgfplotsset{
  set layers,
  mark layer=axis tick labels
}
  \begin{axis}[
    scatter,
    ymax=45,
    ymin=25,
    xmin=25,
    xmax=70,
    xlabel={R@100},
    ylabel={mR@100},
    width=\linewidth,
    height=0.5\linewidth,
    grid=major,
    major grid style={line width=0.2pt, dashed, draw=gray!50},
    legend style={fill=white, fill=none, draw opacity=1, text opacity=1,nodes={scale=1.2, transform shape}, at={(1, 1)}, 
    inner sep=2pt
    },
    legend cell align={left},
    legend columns=1
    ]
    \addlegendimage{mark=*, mark size=2pt, dashed, mark options={draw opacity=0}, draw opacity=1, fill={rgb,255:red,255;green,34;blue,65}}
    \addlegendentry{$\lambda$}

        \addplot[black, dashed, thick, scatter, 
        scatter src=explicit symbolic,
        mark options={draw=none},
        visualization depends on={value \thisrow{Method} \as \methodname},
        scatter/classes={
            {repsgg}={mark=*, mark size=2pt, draw opacity=0, fill=mycolor}
        },
        nodes near coords*={\methodname}, 
        every node near coord/.append style={font=\small, anchor=south, yshift=-15pt, xshift=0pt, color=black}
        ] table[x=r100, y=mr100, meta=backbone, col sep=comma] {r_mr.csv};

    \node[anchor=center, yshift=-17pt, xshift=0pt] at (axis cs:29.06,42.86) {(default)};
    \node[anchor=center, yshift=12pt, xshift=5pt] at (axis cs:64.24,27.12) {no PGLA};
  \end{axis}

\end{tikzpicture}
\caption{Effects of the hyper-parameter $\lambda$ on the trade-off between R@100 and mR@100.}
\label{fig:r_mr}
\end{figure}
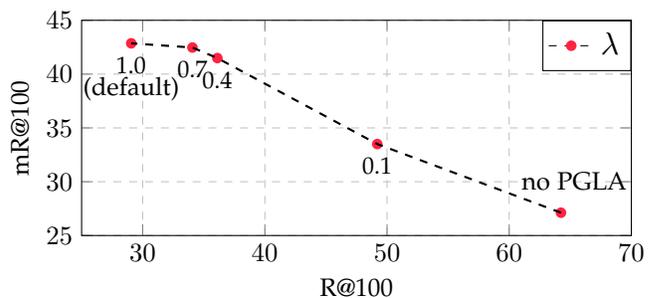

The hyper-parameter $\lambda$ in (\ref{eq:w_b}) allows for adjusting the sensitivity to run-time recall, and we explore its effects on recall and mean recall. 
We evaluate on the PredCls task with different values of $\lambda$, and the results are shown in Fig.~\ref{fig:r_mr}.
As anticipated, increasing $\lambda$ results in higher mR@100 but lower R@100, whereas decreasing $\lambda$ leads to lower mR@100 but higher R@100.
Although better trade-offs may exist, exploring them falls beyond the scope of this paper.

\subsubsection{Analysis of Inference Speed}

Beside performance improvements, another set of benefits of RepSGG are its fast inference speed and flexible inference configurations.
In Section~\ref{sec:reppoint_sampler}, the details of inference are discussed where we sample rep-points within ``$3\sigma$''.
We have the option to perform inference with fewer samples. 
We conduct experiments using samples within $3\sigma$, $2\sigma$, $\sigma$, and just the samples at the means ($\mu$).
We also benchmark several methods discussed in Section~\ref{sec:relatedwork} as comparisons.
We measure the average of frames per second (FPS) over all testing images for all selected models, which is evaluated on a single Nvidia GTX 1080 Ti GPU with a batch size of 1.
The settings from different models follow their original papers and implementations, so the inference speed depends on not only the architectures, but also inference configurations such as image size, mixed precision, \etc.

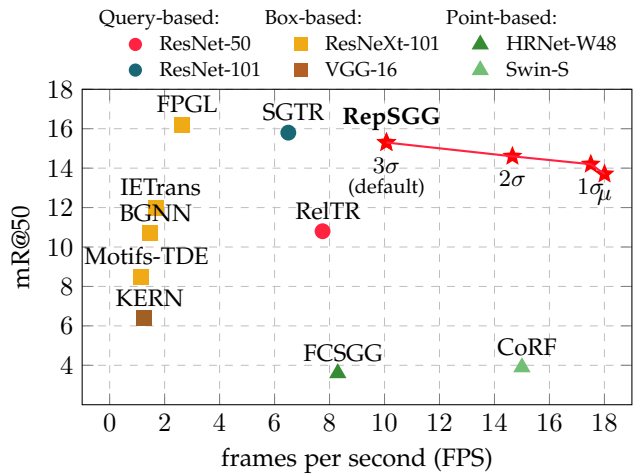
\begin{figure}[t]
\centering
\pgfdeclareplotmark{mystar}{
    \node[star,star point ratio=2.25,minimum size=8pt,
          inner sep=0pt,draw=black,solid,fill=red] {};
}
\definecolor{mycolor}{RGB}{255,34,65}

\begin{tikzpicture}
\pgfplotsset{
  set layers,
  mark layer=axis tick labels
}
  \begin{axis}[
    scatter,
    ymin=2,
    ymax=18,
    xmin=-1,
    xmax=19,
    xtick={0,2,...,18},
    ytick={4,6,...,18},
    xlabel={frames per second (FPS)},
    ylabel={mR@50},
    width=\linewidth,
    height=0.65\linewidth,
    grid=major,
    major grid style={line width=0.2pt, dashed, draw=gray!50},
    legend style={fill=white, draw opacity=1, text opacity=1,nodes={scale=0.8, transform shape}, at={(0.05, 1.01)}, anchor=south west,
    inner sep=1pt,
    column sep=5pt,
    draw=none
    },
    legend cell align={left},
    legend columns=3
    ]
    
    \addlegendimage{empty legend}
    \addlegendentry{\hspace{-1cm}Query-based:}

    \addlegendimage{empty legend}
    \addlegendentry{\hspace{-1cm}Box-based:}

    \addlegendimage{empty legend}
    \addlegendentry{\hspace{-1cm}Point-based:}

    \addlegendimage{only marks, mark=*, mark size=2.5pt, draw opacity=0, fill={rgb,255:red,255;green,34;blue,65}}
    \addlegendentry{ResNet-50}

    \addlegendimage{only marks, mark=square*, mark size=2.5pt, draw opacity=0, fill={rgb,255:red,240;green,170;blue,21}}
    \addlegendentry{ResNeXt-101}

    \addlegendimage{only marks, mark=triangle*, mark size=3.8pt, draw opacity=0, fill={rgb,255:red,54;green,140;blue,50}}
    \addlegendentry{HRNet-W48}

    \addlegendimage{only marks, mark=*, mark size=2.5pt, draw opacity=0, fill={rgb,255:red,25;green,103;blue,116}}
    \addlegendentry{ResNet-101}

    \addlegendimage{only marks, mark=square*, mark size=2.5pt, draw opacity=0, fill={rgb,255:red,180;green,96;blue,36}}
    \addlegendentry{VGG-16}

    \addlegendimage{only marks, mark=triangle*, mark size=3.8pt, draw opacity=0, fill={rgb,255:red,114;green,190;blue,117}}
    \addlegendentry{Swin-S}
    
    \addplot[scatter, 
        only marks, 
        scatter src=explicit symbolic,
        visualization depends on={value \thisrow{Method} \as \methodname},
        scatter/classes={
            {ResNet-50}={mark=*, mark size=3pt, draw opacity=0, fill={rgb,255:red,255;green,34;blue,65}},
            {repsgg}={mark=mystar, mark size=4pt, draw opacity=0, fill={rgb,255:red,255;green,34;blue,65}},
            {ResNet-101}={mark=*, mark size=3pt, draw opacity=0, fill={rgb,255:red,25;green,103;blue,116}},
            {ResNeXt-101}={mark=square*, mark size=3pt, draw opacity=0, fill={rgb,255:red,240;green,170;blue,21}},
            {VGG-16}={mark=square*, mark size=3pt, draw opacity=0, fill={rgb,255:red,160;green,96;blue,36}},
            {HRNet-W48}={mark=triangle*, mark size=3.8pt, draw opacity=0, fill={rgb,255:red,54;green,140;blue,50}},
            {Swin-S}={mark=triangle*, mark size=3.8pt, draw opacity=0, fill={rgb,255:red,114;green,190;blue,117}}
        },
        ] table[x=fps, y=mr50, meta=backbone, col sep=comma] {methodspeed.csv};

        \addplot[
        only marks,
        mark=none,
        visualization depends on={value \thisrow{Method} \as \methodname},
        nodes near coords*={\methodname}, 
        every node near coord/.append style={font=\small, anchor=south, yshift=1.2pt, xshift=2pt}
        ] table[x=fps, y=mr50, meta=backbone, col sep=comma] {methodspeed.csv};

        \addplot[mycolor, thick, scatter, 
        scatter src=explicit symbolic,
        visualization depends on={value \thisrow{Method} \as \methodname},
        scatter/classes={
            {repsgg}={mark=mystar, mark size=4pt, draw opacity=0, fill=mycolor}
        },
        nodes near coords*={\methodname}, 
        every node near coord/.append style={font=\small, anchor=south, yshift=-15pt, xshift=0pt, color=black}
        ] table[x=fps, y=mr50, meta=backbone, col sep=comma] {kspeed.csv};

        \node[anchor=center, yshift=-17pt, xshift=0pt, scale=0.8] at (axis cs:10.07,15.3) {(default)};
  \end{axis}

\end{tikzpicture}
\caption{Inference speed and mR@50 benchmark on the SGDet task.}
\label{fig:method_speed}
\end{figure}

The results of mR@50 and FPS are collected on the SGDet task as shown in Fig.~\ref{fig:method_speed}.
Remarkably, the proposed methods achieve better trade-offs between performance and speed.
No crucial performance drop is observed when less rep-points are sampled.
By sampling within $3\sigma$, our model achieves competitive results compared to the state-of-the-art box-based FPGL and query-based SGTR, while being approximately 4 times and 1.6 times faster, respectively.
By only sampling at the means, RepSGG\textsubscript{PGLA} achieves 13.7 mR@50 with a nearly real-time inference speed of 18 FPS.

\section{Limitations and Future Work} \label{sec:lim}
RepSGG excels on PredCls and SGCls tasks, showcasing an advantage over query-based methods~\cite{li2022sgtr,desai2022single,teng2022structured}.
Nevertheless, its performance on SGDet does not outperform that of the state-of-the-art methods, because it does not fully exploit the ground-truth information as effectively as box-based methods.
Giving the GT bounding boxes, we must map the corner coordinates to appropriate feature levels and identify the best-matched pixels responsible for the detections.
However, the best-matched features do not correspond as accurately to actual ground-truth features as achieved by RoIAlign~\cite{he2017mask}.
A hybrid representation of box, point, and query features could help improve the performance.
Additionally, it is important to note that our model is exploratory and primarily focuses on visual features.
The integration of multi-modal features, such as depth maps~\cite{sharifzadeh2021improving}, language~\cite{lu2016visual}, video~\cite{ji2020action}, and knowledge graph~\cite{zareian2020bridging}, can be seamlessly incorporated into our RepSGG architecture as additional queries and keys.
The architecture also offers potential for extension to other relationship-related tasks, such as human-object interaction (HOI) detection~\cite{chao2018learning}, video-based HOI detection~\cite{chiou2021st}, video SGG~\cite{shang2017video}, panoptic SGG~\cite{yang2022panoptic}, and panoptic video SGG~\cite{yang2023panoptic}.

The proposed PGLA serves as a general approach for leveraging performance evaluation to attain a more balanced performance. 
It can also be adapted for addressing other long-tailed problems (\eg, in visual recognition), utilizing different metrics (such as accuracy), and incorporating other loss functions (such as cross-entropy) with minor adjustments.

\section{Conclusions}\label{sec:conclusions}

We have explored novel representations of entities and relationships for scene graph generation, and introduced a performance-guided logit adjustment strategy for long-tailed learning. 
The proposed RepSGG architecture models entities as subject queries and object keys, and relationships as the attention weights between subjects and objects.
The proposed PGLA significantly mitigates the long-tailed problem in SGG.
Our experiments demonstrate that RepSGG trained with PGLA compares favourably against box-based, query-based, and point-based SGG models with considerably less design complexity.
Our methods also achieve the state-of-the-art performance with fast inference speed.
Due to its effectiveness and efficiency, we envision RepSGG to serve as a strong and simple alternative to current mainstream SGG methods.

\ifCLASSOPTIONcompsoc
  \section*{Acknowledgments}
\else
  \section*{Acknowledgment}
\fi

This work was supported in part by Bourns Endowment funds, National Science Foundation (NSF) awards CNS-1730158, ACI-1540112, ACI-1541349, OAC-1826967, OAC-2112167, CNS-2100237, CNS-2120019, the University of California Office of the President, and the University of California San Diego's California Institute for Telecommunications and Information Technology/Qualcomm Institute. Thanks to CENIC for the 100Gbps networks.

\ifCLASSOPTIONcaptionsoff
  \newpage
\fi



%
\bibliographystyle{IEEEtran}
\bibliography{IEEEabrv,7_abrv_bib}

\end{document}